%% file: main.tex
\documentclass[journal]{IEEEtran}
\usepackage{amsmath,amsfonts}
\usepackage{stfloats}
\usepackage{url}
\usepackage{graphicx}
\usepackage{cite}
\usepackage{booktabs}
\usepackage{multirow}
\usepackage{threeparttable}
\usepackage{pifont}
\usepackage{setspace}
\usepackage{mathrsfs}
\usepackage{enumitem}
\usepackage{float}
\usepackage{ulem}
\usepackage{amssymb}
\usepackage{colortbl}
\usepackage{soul}
\usepackage{makecell} 
\usepackage{hyperref}
\makeatletter
\let\MYcaption\@makecaption
\makeatother
\usepackage[font=footnotesize]{subcaption}
\makeatletter
\let\@makecaption\MYcaption
\makeatother
\hypersetup{
    colorlinks=true,
    linkcolor=blue,
    filecolor=black,      
    urlcolor=black,
    citecolor=purple
}
\input{packages.tex}
\usepackage[ruled,noend]{algorithm2e}
\newcommand\cl[1]{{\color{black}{#1}}}
\newcommand{\settablefont}{\fontsize{6.6}{11.8}\selectfont}

\begin{document}
\normalem

\title{Playing to Vision Foundation Model's\\Strengths in Stereo Matching}

\author{Chuang-Wei Liu$^{\orcidicon{0000-0003-0260-6236}\,}$, Qijun Chen,~\IEEEmembership{Senior Member,~IEEE}, and Rui Fan$^{\orcidicon{0000-0003-2593-6596}\,}$,~\IEEEmembership{Senior Member,~IEEE}
\thanks{
\cl{
This research was supported by the National Natural Science Foundation of China under Grant 62233013, the Science and Technology Commission of Shanghai Municipal under Grant 22511104500, the Fundamental Research Funds for the Central Universities, and Xiaomi Young Talents Program. (\emph{Corresponding author: Rui Fan})}
}
\thanks{
\cl{
Chuang-Wei Liu, Qijun Chen, and Rui Fan are with the College of Electronics \& Information Engineering, Shanghai Research Institute for Intelligent Autonomous Systems, the State Key Laboratory of Intelligent Autonomous Systems, and Frontiers Science Center for Intelligent Autonomous Systems, Tongji University, Shanghai 201804, P. R. China (e-mail: \{cwliu, qjchen, rfan\}@tongji.edu.cn).
}
}
}

%\markboth{IEEE Transactions on Int5elligent Vehicles}%
%{Liu \MakeLowercase{\textit{et al.}}: Playing to Vision Foundation Model's Strengths in Stereo Matching}

\maketitle

\cl{
\begin{abstract}
Stereo matching has become a key technique for 3D environment perception in intelligent vehicles. For a considerable time, convolutional neural networks (CNNs) have remained the mainstream choice for feature extraction in this domain. Nonetheless, there is a growing consensus that the existing paradigm should evolve towards vision foundation models (VFM), particularly those developed based on vision Transformers (ViTs) and pre-trained through self-supervision on extensive, unlabeled datasets. While VFMs are adept at extracting informative, general-purpose visual features, specifically for dense prediction tasks, their performance often lacks in geometric vision tasks. This study serves as the first exploration of a viable approach for adapting VFMs to stereo matching. Our ViT adapter, referred to as ViTAS, is constructed upon three types of modules: spatial differentiation, patch attention fusion, and cross-attention. The first module initializes feature pyramids, while the latter two aggregate stereo and multi-scale contextual information into fine-grained features, respectively. ViTAStereo, which combines ViTAS with cost volume-based stereo matching back-end processes, achieves the top rank on the KITTI Stereo 2012 dataset and outperforms the second-best network StereoBase by approximately 7.9\% in terms of the percentage of error pixels, with a tolerance of 3 pixels. Additional experiments across diverse scenarios further demonstrate its superior generalizability compared to all other state-of-the-art approaches. We believe this new paradigm will pave the way for the next generation of stereo matching networks.
\end{abstract}
}

\section{Introduction} 
\label{sec.intro}

\cl{
\IEEEPARstart{V}{ision} foundation models (VFMs) have rapidly emerged as a focal point in the field of computer vision. From models like Segmentation Anything \cite{kirillov2023segment} and DINOv2 \cite{oquab2023dinov2} by Meta AI to the more recent Depth Anything \cite{yang2024depth}, VFMs have garnered significant attention and interest. Surprisingly, despite its fundamental role in 3D computer vision, stereo matching has not received adequate attention amidst the wave of VFMs yet. Therefore, this paper serves as the first attempt to navigate stereo matching into this new continent, with a specific emphasis on adapting VFMs for more generalizable stereo matching.

Recent advancements in stereo matching have primarily focused on the back-end processes, including cost volume construction \cite{guo2019group}, cost aggregation \cite{xu2023iterative}, and disparity refinement \cite{xu2023unifying}, while relatively overlooking the development of deep feature extractors. This is largely attributed to the effectiveness of traditional backbone networks such as ResNet \cite{he2016deep} and MobileNet \cite{sandler2018mobilenetv2} in extracting rich deep features for matching cost computation. Nevertheless, recent VFMs, generally built upon a Vision Transformer (ViT), have demonstrated greater effectiveness in learning informative, general-purpose deep features across various related computer vision tasks, when pre-trained in a self-supervised fashion on large curated datasets \cite{kirillov2023segment,oquab2023dinov2,yang2024depth}. Therefore, a key focus of this paper lies in the development of adapters to selectively leverage these deep features to improve stereo matching. 

Despite the extensive application of VFMs with task-specific adapters for scene parsing tasks \cite{li2021benchmarking,li2022exploring,chen2023vision}, their utilization in geometric vision tasks, such as stereo matching and optical flow estimation, remains limited \cite{li2022practical,guo2022context,liu2024global}. This is primarily because VFMs trained for image segmentation (pixel-level classification) and monocular depth estimation (pixel-level regression) are not capable of producing features that are sufficiently distinct for similarity measurement in the cost volume construction stage \cite{weinzaepfel2023croco}. The significant domain gap between VFM features and those preferred by geometric vision tasks renders existing VFM adapters infeasible for stereo matching. Therefore, the primary objective in designing our VFM adapter is to further enhance feature distinctiveness, thereby reducing ambiguities in stereo matching.

Additionally, it is noteworthy that there is a potential trend among state-of-the-art (SoTA) networks \cite{weinzaepfel2022croco,weinzaepfel2023croco} to shift away from constructing cost volumes for stereo matching. These networks generally employ a Transformer with an encoder-decoder architecture to aggregate stereo knowledge into features at a single view. These features are then taken as input by a dense prediction Transformer \cite{ranftl2021vision} for disparity regression. We are intrigued by the generalizability of such a network design and have conducted extensive experiments across various public datasets. Regrettably, the comprehensive experimental results suggest a shortfall in its performance on new, unseen data, mainly suffering from the scale ambiguity \cite{eigen2014depth}. This limitation could possibly be attributed to the reduced explainability of disparity estimation without the use of cost volumes. This observation further reinforces our motivation for playing to VFM's strengths by developing an effective adapter to fully exploit the general-purpose deep features for cost volume construction, rather than simply regressing disparities from these features without any interpretability.

Therefore, in this article, we introduce \uline{\textbf{ViT} \textbf{A}dapter for \textbf{S}tereo (\textbf{ViTAS})}, playing to the strengths of VFMs in stereo matching. Our proposed ViTAS incorporates three types of modules: (1) the spatial differentiation module (SDM), which captures multi-scale contextual information by initializing feature pyramids, akin to the studies presented in \cite{lipson2021raft,li2022practical,xu2023unifying,xu2023iterative}, (2) the patch attention fusion module (PAFM), which aggregates multi-scale contextual information into fine-grained features, and (3) the cross-attention module (CAM), which aggregates stereo contextual information into extracted features via cross-view interactions. Notably, our newly developed PAFM employs local patch attention and quasi-global attention, devised in accordance with the pixel-to-patch and squeeze-and-excitation manners, to learn the local and global feature weighting parameters, respectively, significantly reducing computational complexity and memory usage compared to the conventional global attention mechanism \cite{shen2021cfnet,zhou2022self,zhao2022semantic}, which learns these features simultaneously. Combining ViTAS with cost volume-based stereo matching back-end processes yields ViTAStereo, a SoTA, powerful, and highly generalizable stereo matching network. \uline{ViTAStereo achieves \textbf{top ranking} on the KITTI Stereo 2012 dataset} and \uline{\textbf{second-best performance} on the KITTI Stereo 2015 dataset} \cite{geiger2012we}, outperforming StereoBase, the current SoTA stereo matching network, by approximately 5.2-11.3\% in the percentage of error pixels. 

We conclude the contributions of this study as follows:
\begin{itemize}
    \item We introduce ViTAS, marking the first research endeavor to fully exploit the informative, general-purpose features extracted by VFMs for stereo matching. 
    \item We develop a novel, lightweight PAFM that learns local and global feature weighting parameters separately, effectively, and efficiently.
    \item We argue that stereo matching networks relying solely on cross-attention mechanism have limited generalizability, primarily due to the absence of cost volumes.
    \item We conduct extensive experiments to demonstrate the SoTA performance and superior generalizability of ViTAStereo across various public datasets. 
\end{itemize}
}
\cl{
The remainder of this article is structured as follows: related works, including ViT adapters and stereo-matching networks, are presented in Sect. \ref{sec.related}. Sect. \ref{sec.method} details our proposed ViTAS. Comprehensive ablation studies and comparative experiments are presented in Sect. \ref{sec.exp}. Finally, in Sect. \ref{sec.conc}, we summarize the results and provide recommendations for future work.
}

\section{Related Work}
\label{sec.related}

\subsection{ViT Adapters}
\cl{
SoTA VFMs generally utilize a plain ViT as their backbone network. To date, ViT adapters have found widespread application in 2D computer vision tasks. For instance, ViTDet \cite{li2021benchmarking,li2022exploring} enables the plain, non-hierarchical ViT architecture to undergo fine-tuning for object detection without the need for redesigning a hierarchical backbone for pre-training. Similarly, ViT-Adapter \cite{chen2023vision} injects image priors into the ViT using an additional attention path, resulting in superior accuracy in both object detection and semantic/instance segmentation tasks. 
On the other hand, recent research on 3D computer vision problems, such as monocular depth estimation, stereo matching, and optical flow estimation, has explored various training strategies, including contrastive learning and self-supervision, to leverage the ViT architecture \cite{xie2021propagate,wang2021dense}. However, there has been relatively less focus on the development of adapters in this field. It is likely that the deep features utilized for scene understanding tasks are not inherently suitable or compatible with geometric vision tasks. 
Therefore, developing an adapter compatible with the plain ViT architecture for geometric vision tasks is a popular area of research that requires more attention.}
\subsection{Stereo Matching}
\subsubsection{Cost Volume-based Networks}
\cl{
Recent SoTA stereo matching networks \cite{lipson2021raft,li2022practical,xu2023iterative,xu2023unifying} based on cost volumes have largely overlooked pyramid feature extraction and instead focused mainly on back-end stages that process features and costs. Since the introduction of RAFT-Stereo \cite{lipson2021raft}, its core component--a multi-level gated recurrent unit, has become prevalent in stereo matching. This unit takes feature pyramids extracted by a conventional deep feature extractor as input for cost aggregation, enabling the incorporation of semantic information at various scales. RAFT-Stereo has significantly influenced subsequent advancements in stereo matching networks, such as those seen in CREStereo \cite{li2022practical}, IGEV-Stereo \cite{xu2023iterative}, and GMStereo \cite{xu2023unifying}. These networks have further extended and refined this multi-scale structure, often integrating attention mechanism to improve feature representation and address local matching ambiguities in challenging regions. For instance, CREStereo \cite{li2022practical} follows LoFTR \cite{sun2021loftr} and incorporates an attention module at the lowest resolution to aggregate global contextual and stereo information in single or cross-view feature maps. IGEV-Stereo \cite{xu2023iterative}, on the other hand, combines multi-scale correlation volumes with a geometry encoding volume obtained through 3D convolutions, aiming at addressing local matching ambiguities in ill-posed regions. Moreover, GMStereo \cite{xu2023unifying} further extends this multi-scale structure into both stereo matching and optical flow estimation tasks, with the exactly same learnable parameters. Similar to CreStereo, GMStereo also employs attention modules prior to cost volume construction, but across all spatial scales. In contrast to these SoTA methods, we leverage recent VFMs for feature extraction and design our adapter, drawing inspiration from the existing pyramid feature structure and attention mechanism.
}

\cl{
\subsubsection{Cost Volume-free Network} 
CroCo-Stereo \cite{weinzaepfel2023croco} is a seminal contribution in this domain. It utilizes an encoder-decoder Transformer, initially pre-trained on the ImageNet database \cite{deng2009imagenet} for cross-view completion \cite{weinzaepfel2022croco}, and subsequently fine-tuned for stereo matching or optical flow estimation by incorporating a dense prediction Transformer (DPT) head \cite{wang2021dense}. Specifically, stereo information is aggregated into the left-view features through cross-attention via a plain ViT decoder, followed by a DPT head to regress disparities. However, due to the utilization of a large VFM encoder and extensively deployed attention modules in the decoder, CroCo-Stereo incurs even greater computational and memory overhead compared to cost volume-based networks. More importantly, abandoning the cost volume considerably reduces its generalizability and interoperability, as demonstrated by our extensive experiments. These limitations underscore the importance of prioritizing compatibility with cost volume-based networks.
}

\begin{figure*}[t!]
	\begin{center}
		\centering
		\includegraphics[width=0.99\textwidth]{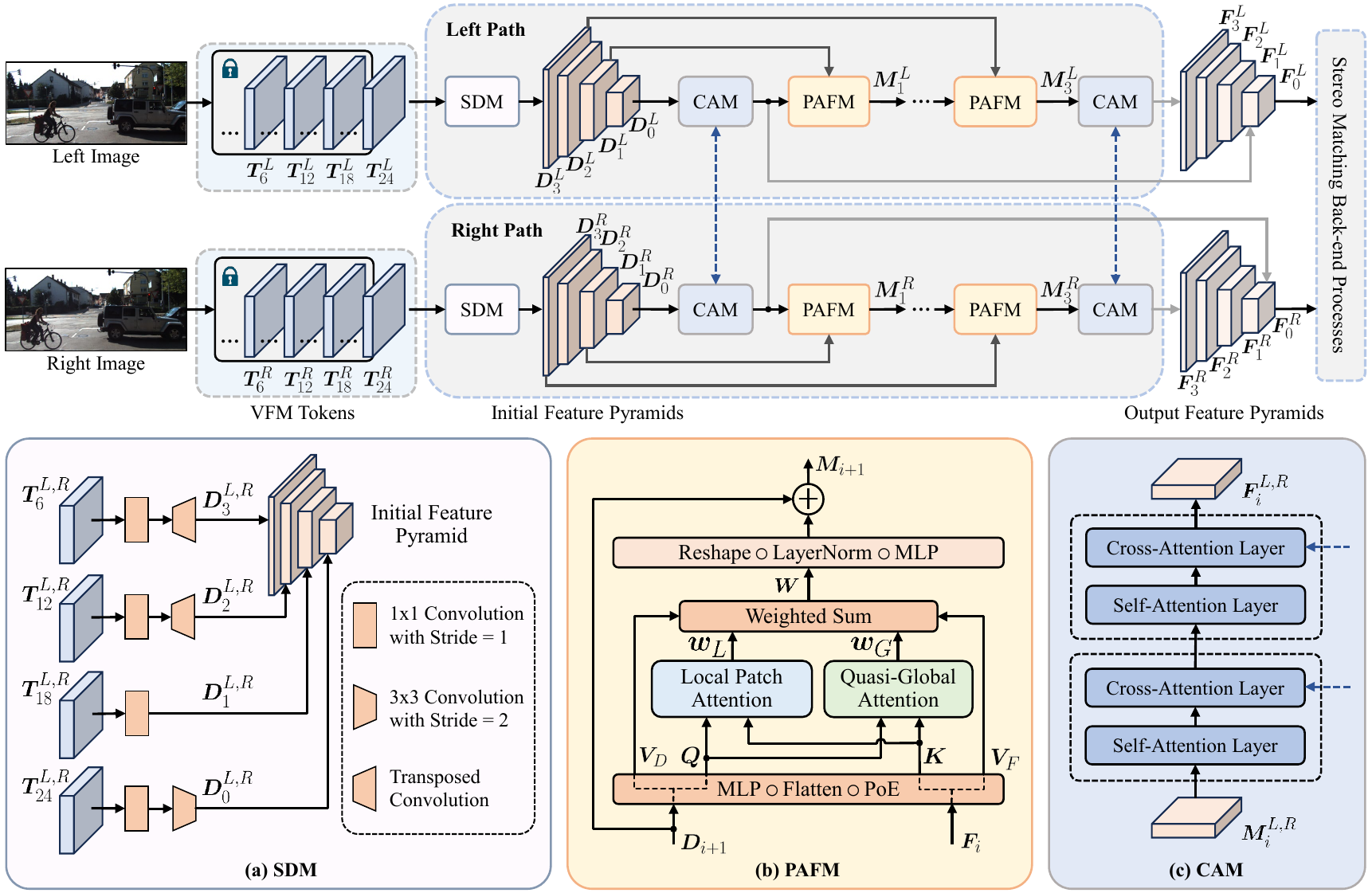}
		\centering
		\caption{\cl{An illustration of our proposed \textbf{ViTAS}, consisting of \textbf{an SDM}, \textbf{four CAMs}, and \textbf{three PAFMs} for each sub-network.}}
		\label{fig.framework}
	\end{center}
\end{figure*}

\section{Methodology}
\label{sec.method}

\cl{The overall task is first formulated in Sect. \ref{sec.task}. Then, the three modules in ViTAS are detailed in Sects. \ref{sec.sdm}, \ref{sec.PAFM}, and \ref{sec.cam}, respectively.
}
\subsection{Task Formulation}
\label{sec.task}

\cl{
Pyramid feature extraction has been prevalently used in stereo matching \cite{lipson2021raft,li2022practical,xu2023unifying}, primarily due to its capability to handle objects of various scales while maintaining high computational efficiency \cite{wang2022pvt,quan2023centralized}. As demonstrated in recent works \cite{lipson2021raft,ma2022multiview,li2022practical,xu2023iterative,xu2023unifying,xu2023accurate}, features at four scales ($1/32$, $1/16$, $1/8$, and $1/4$ of the original image resolution) have been shown to be sufficient and effective for stereo matching.  
However, the SoTA VFMs \cite{kirillov2023segment,oquab2023dinov2} generally adopt the plain ViT architecture \cite{dosovitskiy2020image} for feature extraction, resulting in extracted features at a single resolution. Specifically, a VFM consists of a patch embedding layer and $N$ consecutive Transformer encoders. An input image $\boldsymbol{I} \in \mathbb{R}^{h_0 \times w_0 \times 3}$ is first divided into a collection of $p \times p$ non-overlapping patches (typically $p=16$) by the patch embedding layer. These patches are then sequentially projected into $N$ tokens $\boldsymbol{T} \in \mathbb{R}^{h_0/p \times w_0/p \times c_{T}}$ through the Transformer encoders. 
Therefore, to fully exploit the general-purpose VFM features, the most fundamental task of our ViTAS is to transform the tokens into a collection of pyramid features $\mathcal{F}=\{\boldsymbol{F}_{0},\boldsymbol{F}_{1},\boldsymbol{F}_{2}, \boldsymbol{F}_{3}\}$, where $\boldsymbol{F}_{k}\in\mathbb{R}^{\frac{h_0}{2^{5-k}} \times \frac{w_0}{2^{5-k}} \times {c}_k}$.

In this study, we use DINOv2 \cite{oquab2023dinov2} as our backbone VFM, in which $N$ is set to 24. 
Nonetheless, a limitation arises as it sets the parameter $p$ to 14, causing a misalignment between token scales and the preferred pyramid feature scales. To address this issue, we first adjust the input stereo images by a factor of $\frac{14}{16}$, resulting in tokens at $\frac{1}{16}$ of the original image resolution. Subsequently, following recent advancements in ViT adapters \cite{li2021benchmarking,chen2023vision}, we split the Transformer encoders into four groups. From each group, we select tokens generated by the final Transformer block to serve as input for our ViTAS. Consequently, the essential goal of our ViTAS is to transform the $\mathcal{T}^{L,R}=\{\boldsymbol{T}_{6}^{L,R},\boldsymbol{T}_{12}^{L,R},\boldsymbol{T}_{18}^{L,R},\boldsymbol{T}_{24}^{L,R}\}$ into $\mathcal{F}^{L,R}$, where the superscripts ${L}$ and ${R}$ correspond to the left and right images, respectively.
}

\begin{algorithm}[t!]
\small
\normalem
\setstretch{1.45}
\caption{ViTAS workflow}
\label{alg.adapter}
\LinesNumbered 
\KwIn{VFM tokens $\mathcal{T}^{L,R}$}
\KwOut{Output feature pyramids $\mathcal{F}^{L,R}$}
Generating initialized feature pyramids from VFM tokens via: 
$\mathcal{D}^{L,R} \gets \mathrm{SDM}(\mathcal{T}^{L,R})$\;
Aggregating stereo contextual information for the deepest initial features via:
$\{ \boldsymbol{F}_{0}^{L},\boldsymbol{F}_{0}^{R} \} \gets \text{CAM}(\boldsymbol{D}_0^{L},\boldsymbol{D}_0^{R})$ \;
Aggregating stereo and multi-scale contextual information hierarchically via:
\For {$i \gets 1 \enspace \text{to} \enspace 3$ } {
        $\boldsymbol{M}_{i}^{L,R} \gets \text{PAFM}(\boldsymbol{F}_{i-1}^{L,R},\boldsymbol{D}_{i}^{L,R})$\;
        $\boldsymbol{F}_{i}^{L,R} \gets \text{CAM}(\boldsymbol{M}_i^{L,R},\boldsymbol{M}_i^{R,L})$\;
        }
\end{algorithm}

\subsection{ViTAS Architecture}
\label{sec.adapter}

\cl{As depicted in Fig. \ref{fig.framework}, our proposed ViTAS adopts a twin architecture comprising two weight-sharing sub-networks. Each sub-network is dedicated to processing one view of the stereo images and consists of a SDM followed by three PAFMs and four CAMs arranged alternately. The SDM generates initial feature pyramids, while the PAFMs and CAMs hierarchically aggregate stereo and multi-scale contextual information into fine-grained features, respectively. Each module type is designed to accomplish an independent task, making our ViTAS highly modularized and adaptable to future updates with more advanced techniques. The workflow of our proposed ViTAS is detailed in Algorithm \ref{alg.adapter}. 
}

\subsubsection{SDM}
\label{sec.sdm}

\cl{
Recent studies \cite{park2022vision,fang2023unleashing} have demonstrated the complementarity between ViTs and convolutional neural networks (CNNs). The former excel at capturing global contextual information, while the latter enrich the local spatial patterns of the ViT tokens.
Therefore, we introduce the SDM at the beginning of ViTAS to re-scale the ViT tokens, as illustrated in Fig. \ref{fig.framework}(a). This process enables ViTAS to capture multi-scale contextual information, resulting in significantly improved stereo matching accuracy, particularly for small objects and boundaries. 

The input ViT tokens $\mathcal{T}^{L,R}$ are first assembled into initial feature pyramids $\mathcal{D}^{L,R}=\{\boldsymbol{D}_{0}^{L,R},\boldsymbol{D}_{1}^{L,R},\boldsymbol{D}_{2}^{L,R},\boldsymbol{D}_{3}^{L,R} \}$ using two SDMs, each consisting of four blocks of convolutions and transpose convolutions, where $\boldsymbol{D}_{k}^{L,R}$ is at a resolution equal to $1/2^{5-k}$ of the original image. Since deeper ViT tokens contain richer global context and shallower ones focus on fine-grained details \cite{ranftl2021vision}, we assemble ViT tokens from deeper to shallower layers with gradually increasing resolutions. In addition, we reduce the number of channels in the initial feature pyramids to further alleviate computational and memory pressure. A hierarchical refinement process is then performed with CAMs and PAFMs from $\boldsymbol{D}_{0}^{L,R}$ to $\boldsymbol{D}_{3}^{L,R}$, as detailed in the remainder of this section.
}

\subsubsection{PAFM}
\label{sec.PAFM}

\begin{figure}[t!]
	\begin{center}
		\centering
		\includegraphics[width=0.48\textwidth]{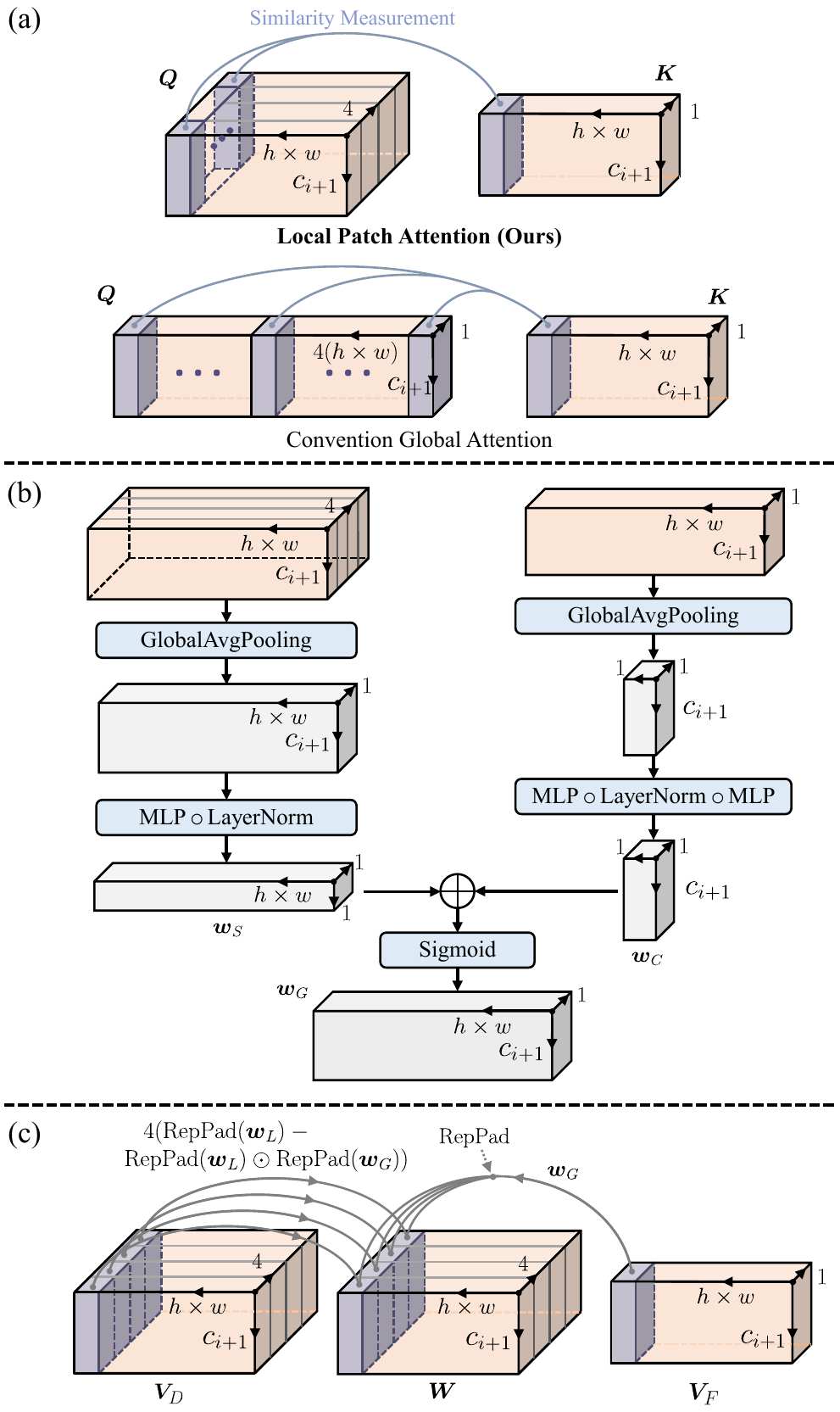}
		\centering
		\caption{\cl{Illustrations of (a) local path attention versus conventional global attention, (b) quasi-global attention, and (c) multi-scale feature aggregation within PAFM.}}
		\label{fig.pafm}
	\end{center}
\end{figure}

With the feature pyramids $\mathcal{D}^{L,R}$ initialized by SDM, we perform multi-scale feature fusion to aggregate $\boldsymbol{F}_i$ and  $\boldsymbol{D}_{i+1}$. Although recent Transformer-based multi-scale feature fusion approaches \cite{zhao2022semantic,wang2023cross} have demonstrated superior performance over CNN-based methods \cite{fan2020sne,chang2020attention,shen2021cfnet,zhou2022self} in dense prediction tasks \cite{zhou2022self}, their computational complexity and memory consumption, inherent in the global attention mechanism, pose significant demands.
Furthermore, a common limitation persists wherein lower-resolution feature maps have to be upsampled (typically via bilinear interpolation) to align with the higher-resolution feature maps. However, such simplistic feature upsampling operations fail to preserve fine-grained details in low-resolution features \cite{liu2023learning,liu2023devil}. 
To overcome these limitations, we design PAFM, a lightweight yet effective Transformer-based multi-scale feature fusion module. 
As illustrated in Fig. \ref{fig.framework}(b), our proposed PAFM consists of a local patch attention and a quasi-global attention, capable of learning local weights $\boldsymbol{w}_{L} \in \mathbb{R}^{(h \times w) \times 4 \times 1}$ and global weights $\boldsymbol{w}_{G}\in \mathbb{R}^{(h \times w) \times 1\times c_{i+1}}$, respectively. $\boldsymbol{F}_{i}$ is aggregated based on $\boldsymbol{w}_{G}$, while $\boldsymbol{D}_{i+1}$ is aggregated based on both $\boldsymbol{w}_{G}$ and $\boldsymbol{w}_{L}$.

The local patch attention measures the fine-grained feature similarity between a given pixel in $\boldsymbol{F}_{i} \in \mathbb{R}^{h \times w \times c_i}$ and its corresponding patch with a resolution of $2 \times 2$ pixels in $\boldsymbol{D}_{i+1} \in \mathbb{R}^{2h\times 2w\times c_{i+1}}$. To this end, we project $\boldsymbol{D}_{i+1}$ into query $\boldsymbol{Q} \in \mathbb{R}^{(h \times w) \times 4\times c_{i+1}}$ and value $\boldsymbol{V}_{D} \in \mathbb{R}^{(h \times w) \times 4\times c_{i+1}}$, and project $\boldsymbol{F}_{i}$ into key $\boldsymbol{K} \in \mathbb{R}^{(h \times w) \times 1\times c_{i+1}}$ and another value $\boldsymbol{V}_{F} \in \mathbb{R}^{(h \times w) \times 1\times c_{i+1}}$, where the second dimension, referred to as the ``patch dimension'' in this article, depicts the number of pixels inside a feature patch. Compared to conventional global attention, our local patch attention operates by measuring feature similarity in a pixel-to-patch manner, as illustrated in Fig. \ref{fig.pafm}(a). This approach dramatically reduces the computational demands, lowering the complexity from $\mathcal{O}(N^2)$ to $\mathcal{O}(N)$. The local weights $\boldsymbol{w}_{L}$ are computed through the following process: 
\begin{equation}
\boldsymbol{w}_{L} = \mathrm{Softmax}(\frac{\boldsymbol{Q} \odot \mathrm{RepPad}(\boldsymbol{K}) \times \boldsymbol{O}}{\sqrt{c_{i+1}}}),
\label{eq.local}
\end{equation}
where $\odot$ denotes the element-wise multiplication operation, $\mathrm{RepPad}$ denotes replication padding operation, $\boldsymbol{O} \in \mathbb{R}^{c_{i+1} \times 1}$ is a matrix storing ones, and the $\mathrm{Softmax}$ operation is performed at the patch dimension to normalize the weights within a feature patch. 

The quasi-global attention emphasizes both the informative spatial areas in $\boldsymbol{D}_{i+1}$ and the prominent feature channels in $\boldsymbol{F}_{i}$, yielding the global weights $\boldsymbol{w}_{G}$, as illustrated in Fig. \ref{fig.pafm}(b). 
To this end, we first employ a global average pooling layer to squeeze $\boldsymbol{Q}$ along the patch dimension, thereby aligning its size with $\boldsymbol{K}$. We then aggregate the squeezed features along the channel dimension using a multi-layer perceptron (MLP) to prioritize informative spatial areas while suppressing redundant ones, thereby producing the spatial weights $\boldsymbol{w}_{S} \in \mathbb{R}^{(h \times w)\times 1\times 1}$. In the meantime, we follow the design of squeeze-and-excitation block \cite{hu2018squeeze,gao2022doubly} and produce the context weights $\boldsymbol{w}_{C} \in \mathbb{R}^{1 \times 1\times c_{i+1}}$ from $\boldsymbol{K}$ (containing rich global contextual information), highlighting prominent feature channels while de-emphasizing the less important ones. This is achieved through a combination of a global average pooling layer and two MLPs. Finally, the global weights $\boldsymbol{w}_{G}$ are calculated as follows:
\begin{equation}
\boldsymbol{w}_{G} = \mathrm{Sigmoid}\left(\mathrm{RepPad}\left(\boldsymbol{w}_{C}\right) \oplus \mathrm{RepPad}\left(\boldsymbol{w}_{S}\right)\right),
\label{eq.global}
\end{equation}
where $\oplus$ denotes the element-wise summation operation. Afterwards, as depicted in Fig. \ref{fig.pafm}(c), we combine $\boldsymbol{w}_{L}$ and $\boldsymbol{w}_{G}$, the local and global weights, to adaptively fuse $\boldsymbol{V}_{D}$ and $\boldsymbol{V}_{F}$ as follows:
\begin{equation}
\begin{aligned}
&\boldsymbol{W} = \mathrm{RepPad}\left(\boldsymbol{w}_G \odot \boldsymbol{V}_{F} \right) \oplus
\\
&4\left(\mathrm{RepPad}\left(\boldsymbol{w}_{L}\right)-\mathrm{RepPad}\left( \boldsymbol{w}_{L}\right)\odot\mathrm{RepPad}\left(\boldsymbol{w}_{G}\right)
\right) \odot \boldsymbol{V}_{D}),
\end{aligned}
\label{eq.fuse}
\end{equation}
where $\boldsymbol{W} \in \mathbb{R}^{(h \times w) \times 4\times c_{i+1}}$ denotes the fused features. 
Consequently, multi-scale contextual information is aggregated into $\boldsymbol{W}$ with weight parameters balancing between $\boldsymbol{Q}$ and $\boldsymbol{V}$ (summing to 4 in each feature patch). With fully fused information from $\boldsymbol{D}_{i+1}$ and $\boldsymbol{F}_{i}$, the final output of our PAFM is derived as follows:
\begin{equation}
\boldsymbol{M}_{i+1} = \boldsymbol{D}_{i+1} + \mathrm{Reshape} \circ \mathrm{LayerNorm} \circ \mathrm{MLP} (\boldsymbol{W}),
\label{eq.res}
\end{equation}
which is subsequently fed into a CAM for stereo contextual information aggregation.

\subsubsection{CAM}
\label{sec.cam}
\cl{
CAMs have been commonly utilized in stereo matching networks \cite{li2022practical,su2022chitransformer,xu2023unifying} to aggregate stereo contextual information into features in both views through cross-view feature interactions. In this study, CAMs are strategically positioned after the SDM and are interleaved with PAFMs, prior to the generation of the output feature pyramids $\mathcal{F}^{L,R}$. As depicted in Fig. \ref{fig.framework}(c), each CAM contains two attention blocks, each of which consists of a self-attention layer and a cross-attention layer, respectively. The former aggregates global contextual information, whereas the latter enhances feature distinctiveness, thereby reducing disparity ambiguities, particularly in texture-less and occluded regions. Despite having similar structures and sharing the same query feature sources, these two layers diverge in key and value feature sources: the self-attention layer uses features from the same view, whereas the cross-attention layer uses features from the other view. The existing networks for stereo matching suffer from a crucial limitation due to the absence of cross-scale feature interaction, leading to an excessive dependency on self-attention layers to capture global contextual information. To address this issue, GMStereo \cite{xu2023unifying} utilizes six attention blocks within each CAM to independently process features across different layers, albeit at a notable increase in computational complexity. In contrast, our proposed ViTAS has a progressive architecture, wherein PAFMs collaborate with CAMs to aggregate both global and stereo contextual information from deeper layers into shallower ones. As a result, ViTAS utilizes a markedly more lightweight CAM to process features, significantly reducing the computational complexity and memory demands in comparison to GMStereo.
}

\section{Experiments}
\label{sec.exp}

\cl{
This section comprehensively analyzes the effectiveness of our proposed ViTAS in improving both disparity accuracy and network generalizability.} 
\cl{The following subsections delve into details on datasets and implementations, evaluation metrics, ablation studies, and a thorough performance evaluation.
}
 
\subsection{Datasets and Implementation Details}
\cl{Five public stereo matching datasets are utilized in our experiments for model pre-training and fine-tuning. The following two synthetic, large-scale datasets with dense disparity ground truth are employed for the first purpose:}
\begin{enumerate}
    \item \cl{\textbf{SceneFlow} \cite{mayer2016large} consists of a training set (containing 35,454 stereo image pairs) and a test set (often known as the Flying 3D test set, containing 4,370 stereo image pairs). We use the ``finalpass'' version rather than the ``cleanpass'' version because it is more realistic.}
    \item \cl{\textbf{Virtual KITTI} \cite{mayer2016large} contains 21,260 stereo image pairs, generated from five different virtual worlds (created using the Unity game engine and a real-to-virtual cloning method) in urban settings under different imaging and weather conditions.} 
\end{enumerate}
\cl{The following three real-world, small datasets are used to fine-tune networks and evaluate their performance:}
\begin{enumerate}
    \cl{\item \textbf{KITTI Stereo} contains two subsets: 2012 \cite{geiger2012we} and 2015 \cite{menze2015object}, with 192 and 200 training pairs, respectively, and 194 and 200 test pairs, respectively. Sparse disparity ground truth is generated using a LIDAR. 
    \item \textbf{Middlebury} \cite{scharstein2007learning,hirschmuller2007evaluation,scharstein2014high} contains five subsets: 2005, 2006, 2014, 2021, and MiddEval3, with 45, 171, 132, 335, and 14 pairs of high-resolution indoor stereo images and their corresponding disparity ground truth provided by structured light, respectively.  
    \item \textbf{ETH3D} \cite{schops2017multi} contains 27 pairs of stereo grayscale images of both indoor and outdoor scenes. A high-precision laser scanner is used to provide the disparity ground truth.}
\end{enumerate}
\cl{
In our ablation studies (see Sect. \ref{sec.abla}), each network is first trained on the SceneFlow training set and the complete Virtual KITTI dataset for 50 epochs. The pre-trained networks are determined based on their performance on the Flying 3D test set, with the networks demonstrating the best performance being selected. }

\cl{
When comparing our proposed ViTAStereo with SoTA networks on the KITTI test set (see Sect. \ref{sec.sota}), we first pre-train our network on the combined training set of the aforementioned five datasets for 100 epochs. Subsequently, an additional fine-tuning stage is conducted on the KITTI training set for 200 epochs.
}

\cl{
When evaluating the generalizability of our proposed ViTAStereo (see Sect. \ref{sec.cross}), we adopt the same pre-training strategy used in the ablation studies. Specifically, we split the original KITTI training set into two subsets: \textbf{KITTI Train} and \textbf{KITTI Eval}, for model fine-tuning and generalizability evaluation, respectively. Similarly, we divide the original Middlebury dataset into two subsets: \textbf{Midd Train} (excluding the MiddEval3 dataset) and \textbf{Midd Eval} (identical to the MiddEval3 dataset) for model fine-tuning and generalizability evaluation, respectively. The entire \textbf{ETH3D} dataset is used only for generalizability evaluation.
}

\cl{
All experiments are conducted on four NVIDIA RTX 4090 GPUs. During model training, we randomly crop images to $320 \times 720$ pixels and apply conventional data augmentation techniques, including random changes in image color, random rescaling, and random erasing, to further enhance model performance. The back-end components in IGEV-Stereo \cite{xu2023iterative} are used to build our ViTAStereo, primarily due to the similar hardware requirement (both are capable of training and testing on a GPU with 24 GB GDDR6X memory). The parameters of the VFM, excluding those of the last six ViT encoder blocks, are frozen. In our generalizability evaluation experiments, we further demonstrate the compatibility of our proposed ViTAStereo with three additional SoTA stereo matching networks, GMStereo \cite{xu2023unifying}, CREStereo \cite{li2022practical} and CroCo-Stereo \cite{weinzaepfel2023croco}.
% two additional SoTA stereo matching networks, GMStereo \cite{xu2023unifying} and CroCo-Stereo \cite{weinzaepfel2023croco}, are employed to further demonstrate the compatibility of our proposed ViTAStereo with ViT-based networks, especially those incorporating cost volumes. 
The loss function, learning rate, and optimizer used in our experiments are identical to the settings of the reported in their publications \cite{li2022practical,xu2023unifying,xu2023iterative,weinzaepfel2023croco}.
}

\subsection{Evaluation Metrics}

\cl{The following three metrics are computed to quantify stereo matching accuracy (lower values indicate better performance):
\begin{itemize}
    \item \textbf{end-point error (EPE)}, indicating the average disparity estimation error;
    \item \textbf{percentage of error pixels (PEP)}, indicating the percentage of incorrect disparities with respect to a tolerance of $\delta$ pixels;
    \item \textbf{D1}, indicating the percentage of disparities for which the estimation error exceeds both three pixels and 5\% of the ground-truth disparity.
\end{itemize}
}

\begin{figure}[t!]
	\begin{center}
		\centering
		\includegraphics[width=0.49\textwidth]{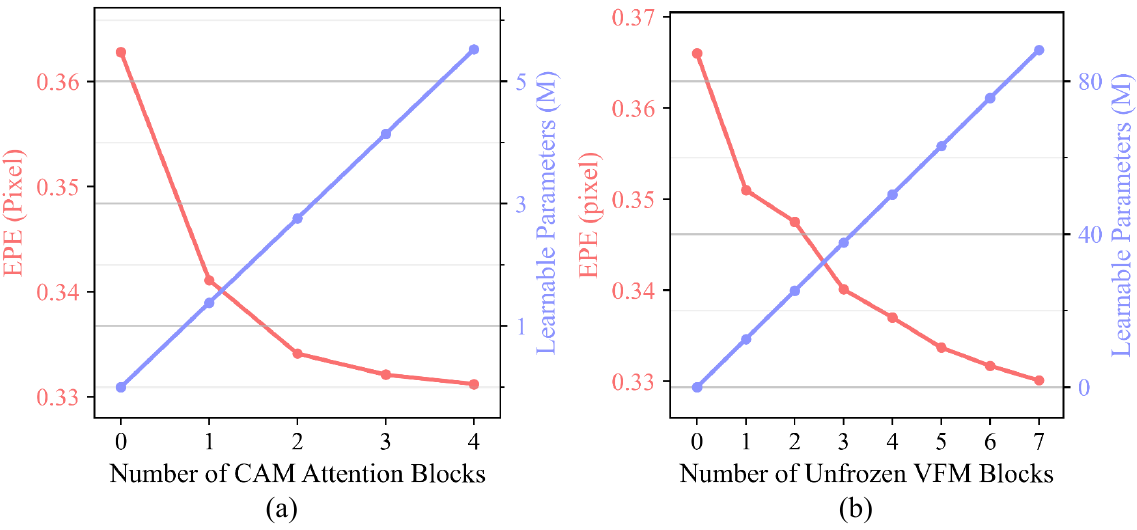}
		\centering
		\caption{Ablation studies on (a) the optimal configuration for CAM attention blocks and (b) the most suitable number of unfrozen VFM blocks.}
		\label{fig.abla}
	\end{center}
\end{figure}

\begin{table}[t!]
    \settablefont
    \caption{Ablation study on the effectiveness of each component within ViTAS. The best results are shown in bold type.}
    \centering
    \label{tab.abla1}
    {
    \setlength{\tabcolsep}{1mm}{
	\begin{tabular}
		{cccccccc}
		\toprule[1.5pt]
		\multirow{2}{*}{SDM} & \multicolumn{3}{c}{Fusion Methods} & \multirow{2}{*}{CAM} & \multirow{2}{*}{EPE (pixel)} & \multirow{2}{*}{D1-all (\%)} & \multirow{2}{*}{Runtime (s)} \\ 
		\cmidrule(r){2-4}
            \specialrule{0em}{-1pt}{-1pt}
            & PAFM & SDFA \cite{zhou2022self} & VAF \cite{zhao2022semantic} \\
		\midrule[1pt]
            \ding{52}&\ding{52}&&&\ding{52}& \textbf{0.334} & \textbf{1.109} & 0.278 \\
            &\ding{52}&&&\ding{52}& 0.351 & 1.206 & 0.275 \\
            \ding{52}&&&&\ding{52}& 0.388 & 1.349 & 0.273 \\
            \ding{52}&\ding{52}&&&& 0.362 & 1.281 & 0.266 \\
            \ding{52}&&&&& 0.427 & 1.389  & 0.262 \\
            &\ding{52}&&&& 0.383 & 1.266 & 0.265 \\
            &&&&\ding{52}& 0.417 & 1.348 & 0.272 \\
            &&&&& 0.435 & 1.394 & \textbf{0.261}  \\
		\midrule
            \ding{52}&&\ding{52}&&\ding{52}& 0.351 & 1.155 & 0.274 \\
            \ding{52}&&&\ding{52}&\ding{52}& 0.335 & 1.122 & 0.287 \\
    	\bottomrule [1.5pt]   	
	\end{tabular}}}
\end{table}

\begin{table}[t!]
    \settablefont
    \caption{The amounts of learnable parameters and memory demands of PAFM and another two meticulously designed feature fusion approaches.}
    \centering
    \label{tab.fuse}
    {
    \setlength{\tabcolsep}{1.5mm}{
	\begin{tabular}
		{ccc}
		\toprule[1.5pt]
		Feature Fusion Approache & Parameters (M) & Memory (MB)\\ 
		\midrule[1pt]
            \textbf{PAFM} & \textbf{0.22} & {118} \\
            SDFA \cite{zhou2022self} & 1.38 & \textbf{98.3} \\
            VAF \cite{zhao2022semantic} & 1.31 & 360 \\
    	\bottomrule [1.5pt]   	
	\end{tabular}}}
\end{table}

\subsection{Ablation Study}
\label{sec.abla}
\cl{
We first investigate the optimal configuration for CAM and determine the most suitable number of unfrozen VFM blocks, as detailed in Fig. \ref{fig.abla}. It is evident that incorporating a greater number of attention blocks and unfreezing additional VFM blocks both contribute to reductions in EPEs, albeit at the cost of a significant increase in the model's learnable parameters. Therefore, in subsequent experiments, we build our CAM with only two attention blocks and unfreeze the last five VFM blocks, so as to minimize the trade-off between disparity accuracy and network complexity. 

In an additional ablation study conducted to validate the effectiveness of each component within our ViTAS, the findings, as detailed in Table \ref{tab.abla1}, demonstrate that the inclusion of any single module leads to improved disparity accuracy. Specifically, the incorporation of SDM, PAFM, and CAM independently leads to reductions in the EPE by 1.8\%, 11.9\%, and 4.1\%, respectively. When all three modules are incorporated, ViTAS achieves the most significant decrease in EPE. This investigation further indicates that excluding any single module from the complete ViTAS yields a performance deterioration comparable to the impact observed when the module is used in isolation. This observation underscores the modular independence within our ViTAS. Notably, the PAFM is identified as the most influential component, primarily attributed to its capability of aggregating both global and stereo contextual information throughout different feature layers, thereby underlining its significance in enhancing the model's overall performance.

Moreover, we compare our PAFM with two other meticulously designed multi-scale feature fusion methods: the CNN-based self-distilled feature aggregation (SDFA) \cite{zhou2022self} and the Transformer-based vertical attention fusion (VAF), to underscore the efficacy of PAFM. As shown in Table \ref{tab.fuse}, PAFM dramatically reduces the number of learnable parameters by 84.1\% and 83.2\% compared to SDFA and VAF, respectively. While PAFM has marginally higher memory requirements than SDFA, its memory demands are significantly lower—by 32.8\%—than those of VAF. Table \ref{tab.abla1} further illustrates that ViTAStereo, equipped with PAFM, outperforms another two feature fusion methods in disparity estimation accuracy. Furthermore, compared to SDFA and VAF, PAFM results in a processing time increase of 1.5\% and a decrease of 3.1\%, respectively. These comprehensive experiments collectively demonstrate that PAFM achieves not only the highest disparity accuracy but also the fewest learnable parameters. In addition, it markedly reduces both computational complexity and memory consumption compared to VAF, demonstrating its exceptional capacity to minimize the trade-off between accuracy and efficiency.
}

\begin{table*}[t!]
    \settablefont
    \caption{Comparison with SoTA stereo matching networks published on the KITTI Stereo 2012 dataset \cite{geiger2012we}. ``$\delta$-noc'' denotes PEP for non-occluded pixels w.r.t. $\delta$, and ``all'' denotes PEP for all pixels w.r.t. $\delta$. The best results are shown in bold type.}
    \centering
    \label{tab.benchmark}
    {
    \setlength{\tabcolsep}{1mm}{
	\begin{tabular}
		{lccccccccc}
		\toprule[1.5pt]
		\multirow{2}{*}{Network} & \multicolumn{8}{c}{PEP w.r.t Different $\delta$} \\ % & \multirow{2}{*}{Ranking} \\
		\cmidrule(r){2-9}
		\specialrule{0em}{-1pt}{-1pt}
		& 2-noc (\%) & 2-all (\%) & 3-noc (\%) & 3-all (\%) & 4-noc (\%) & 4-all (\%)& 5-noc (\%) & 5-all (\%) \\ 
		\midrule[1pt]
            LEAStereo \cite{cheng2020hierarchical} & 1.90 & 2.39 & 1.13 & 1.45 & 0.83 & 1.08 & 0.67 & 0.88 \\
            HITNet \cite{tankovich2021hitnet} & 2.00 & 2.65 & 1.41 & 1.89 & 1.14 & 1.53 & 0.96 & 1.29 \\
            % RAFT-Stereo \cite{lipson2021raft} & 1.92 & 2.42 & 1.30 & 1.66 & 1.30 & 1.66 & & & \\
            ACVNet \cite{xu2022attention} & 1.83 & 2.34 & 1.13 & 1.47 & 0.86 & 1.12 & 0.71 & 0.91 \\
            CREStereo \cite{li2022practical} & 1.72 & 2.18 & 1.14 & 1.46 & 0.90 & 1.14 & 0.76 & 0.95 \\
            PCWNet \cite{shen2022pcw} & 1.69 & 2.18 & 1.04 & 1.37 & 0.78 & 1.01 & 0.63 & 0.81 \\
            IGEV-Stereo \cite{xu2023unifying} & {1.71} & {2.17} & {1.12} & {1.44} & 0.88 & 1.12 & 0.73 & 0.94 \\
            UCFNet \cite{shen2023digging} & 1.67 & 2.17 & 1.09 & 1.45 & 0.85 & 1.12 & 0.69 & 0.91 \\
            ICVP \cite{kwon2023image} & 1.72 & 2.21 & 1.06 & 1.39 & 0.80 & 1.05 & 0.66 & 0.86 \\
            MC-Stereo \cite{feng2023mc} & 1.55 & 1.99 & 1.04 & 1.35 & 0.82 & 1.05 & 0.68 & 0.87 \\
            NMRF-Stereo \cite{guan2024neural} & 1.59 & 2.07 & 1.01 & 1.35 & 0.78 & 1.03 & 0.64 & 0.84 \\
            StereoBase \cite{guo2023openstereo} & 1.54 & 1.95 & 1.00 & 1.26 & 0.76 & 0.97 & 0.62 & 0.80 \\
            \textbf{ViTAStereo (ours)}& \textbf{1.46} & \textbf{1.80} & \textbf{0.93} & \textbf{1.16} & \textbf{0.71} & \textbf{0.87} & \textbf{0.58} & \textbf{0.71} \\
    	\bottomrule [1.5pt]   	
	\end{tabular}}}
    \label{tab.kitti}
\end{table*}

\begin{table}[t!]
    \settablefont
    \caption{Comparison with SoTA stereo matching networks published on the KITTI Stereo 2015 dataset \cite{menze2015object}. D1-bg, D1-fg, and D1-all denote D1 for background, foreground, and all pixels, respectively. }
    \centering
    \label{tab.kitti15}
    {
    \setlength{\tabcolsep}{0.75mm}{
	\begin{tabular}
		{lcccccc}
		\toprule[1.5pt]
		\multirow{2}{*}{Network} & \multicolumn{3}{c}{All Pixels} & \multicolumn{3}{c}{Non-Occluded Pixels} \\
		% \specialrule{0em}{-1pt}{-1pt}
		\cmidrule(r){2-4}
		\cmidrule(r){5-7}
		\specialrule{0em}{-1pt}{-1pt}
		& D1-bg (\%) & D1-fg (\%) & D1-all (\%) & D1-bg (\%) & D1-fg (\%) &D1-all (\%)  \\ 
		\midrule[1pt]
            LEAStereo \cite{cheng2020hierarchical} & 1.40 & 2.91 & 1.65 & 1.29 & 2.65 & 1.51 \\
            HITNet \cite{tankovich2021hitnet} &1.74 & 3.20 & 1.98 & 1.54 & 2.72 & 1.74 \\
            % RAFT-Stereo \cite{lipson2021raft} & 1.58 & 3.05 & 1.82 & & & \\
            CREStereo \cite{li2022practical} & 1.45 & {2.86} & 1.69 & 1.33 & 2.60 & 1.54 \\
            ACVNet \cite{xu2022attention} & {1.37} & 3.07 & 1.65 & 1.26 & 2.84 & 1.52 \\
            UCFNet \cite{shen2023digging} & 1.57 & 3.33 & 1.86 & 1.41 & 2.93 & 1.66 \\
            GMStereo \cite{xu2023unifying} & 1.49 & 3.14 & 1.77 & 1.34 & 2.97 & 1.61 \\
            CroCo-Stereo \cite{weinzaepfel2023croco} & {1.38} & {2.65} & {1.59 }& 1.30 & 2.56 & 1.51 \\
            IGEV-Stereo \cite{xu2023unifying} & {1.38} & {2.67} & {1.59} & 1.27 & 2.62 & 1.49 \\
            MC-Stereo \cite{feng2023mc} & 1.36 & 2.51 & 1.55 & 1.24 & 2.55 & 1.46 \\
            NMRF-Stereo \cite{guan2024neural} & 1.28 & 3.13 & 1.59 & 1.17 & 2.95 & 1.46 \\
            StereoBase \cite{guo2023openstereo} & 1.28 & \textbf{2.26} & \textbf{1.44} & 1.17 & \textbf{2.23} & \textbf{1.35} \\
            % \textbf{IGEV-Stereo (Re)} \cite{xu2023unifying} & 1.89 & 2.40 & 1.26 & 1.57 & 1.55 & 2.91 & 1.78 \\
            \textbf{ViTAStereo (ours)} & \textbf{1.21} & 2.99 & 1.50 & \textbf{1.12} & 2.90 & 1.41 \\
    	\bottomrule [1.5pt]   	
	\end{tabular}}}
\end{table}

\begin{figure*}[t!]
    \centering
    \includegraphics[width=0.99\textwidth]{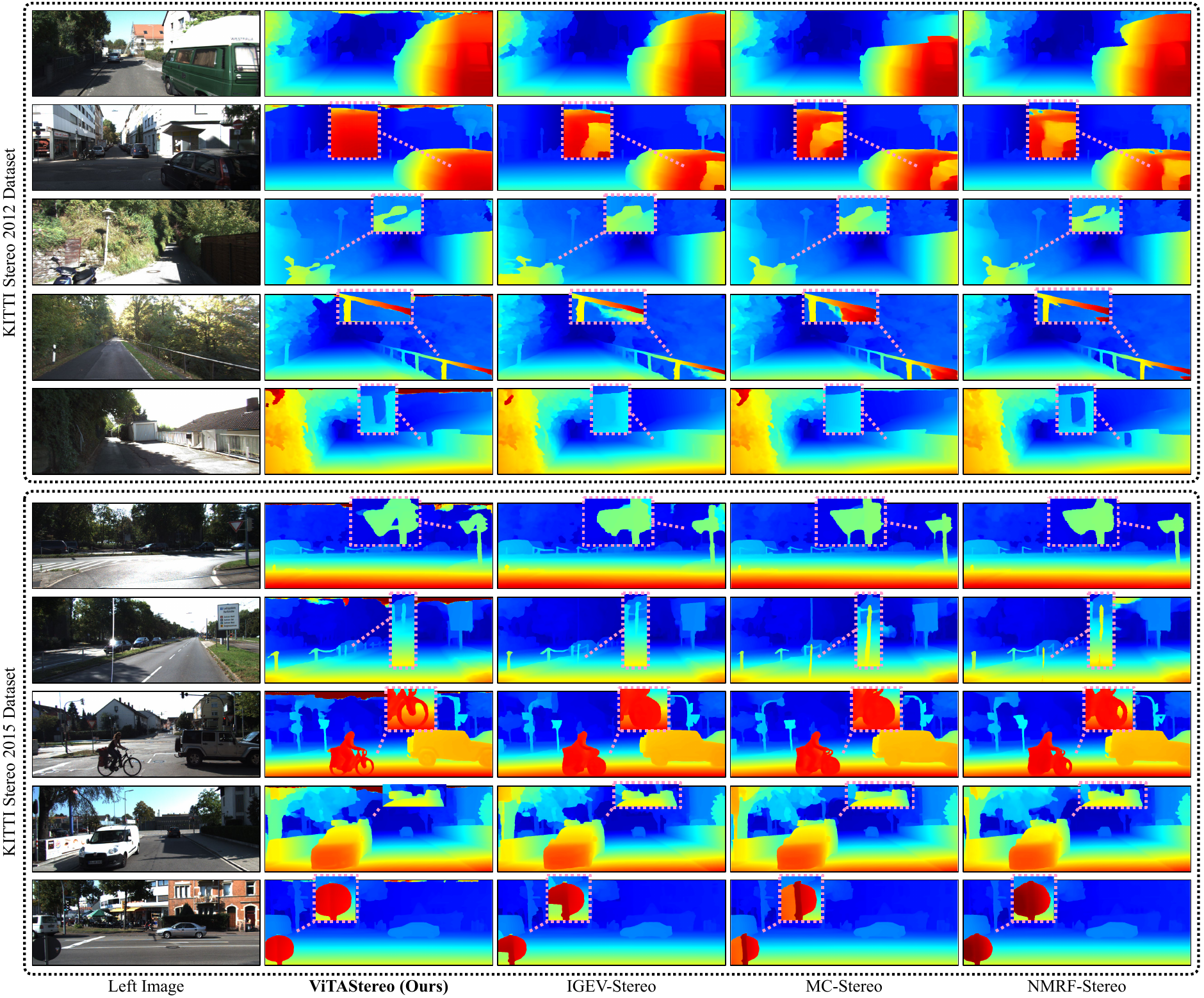}
    \centering
    \caption{Qualitative experimental results on the KITTI Stereo datasets \cite{geiger2012we,menze2015object}, where significantly improved regions are shown with pink dashed boxes.}
    \label{fig.kitti_sota}
\end{figure*}

\subsection{Comparisons with SoTA Networks}
\label{sec.sota}

\cl{
Upon submitting our best results\footnote{These results can be accessed at \url{https://cvlibs.net/datasets/kitti/eval_stereo_flow.php?benchmark=stereo} and \url{https://cvlibs.net/datasets/kitti/eval_scene_flow.php?benchmark=stereo}.} 
% Upon submitting our best results\footnote{These results can be accessed at \href{cvlibs.net/datasets/kitti/eval_stereo_flow.php?benchmark=stereo}{cvlibs.net/datasets/kitti/eval\_stereo\_flow.php?benchmark=stereo} and \href{cvlibs.net/datasets/kitti/eval_scene_flow.php?benchmark=stereo}{cvlibs.net/datasets/kitti/eval\_scene\_flow.php?benchmark=stereo}.} 
(achieved without extensive hyperparameter tuning) to the KITTI Stereo 2012 and 2015 benchmark suites, we conduct a comparative analysis with other SoTA stereo matching networks published on these benchmarks. The results presented in Table \ref{tab.kitti} suggest that on the KITTI Stereo 2012 dataset, our ViTAStereo outperforms all other SoTA stereo matching networks regarding all the evaluation metrics. Surprisingly, ViTAStereo outperforms StereoBase \cite{guo2023openstereo}, the second-best network, by up to 7.00\% and 11.25\% in PEP for non-occluded pixels and all pixels, respectively. Moreover, compared to IGEV-Stereo \cite{xu2023iterative}, which uses an identical stereo matching back-end structure, our ViTAStereo reduces PEP by up to 24.47\%. These significant performance gains underscore the effectiveness of adapting a pre-trained VFM to stereo matching using ViTAS, as opposed to traditional CNN-based backbones for feature extraction. Furthermore, the results presented in Table \ref{tab.kitti15} indicate that ViTAStereo achieves the best performance in terms of D1 in background areas (lower by approximately 5.47\% compared to StereoBase and about 12.3\% compared to IGEV-Stereo) on the KITTI Stereo 2015 dataset. Moreover, our ViTAStereo decreases D1-all by around 5.66\% compared to IGEV-Stereo.

The qualitative experimental results on these two datasets, as illustrated in Fig. \ref{fig.kitti_sota}, also suggest that ViTAStereo outperforms other SoTA networks in handling challenging scenarios. This superior performance is evident in both large-scale, texture-less regions (illustrated in rows 1 and 2) as well as small-scale areas rich in details (shown in rows 3 to 9). We attribute these improvements to the multi-scale feature aggregation process performed by our proposed PAFM. Additionally, ViTAStereo achieves improved disparity accuracy within occluded areas (shown in row 10), further validating the robustness of our approach in complex environments.
}

\begin{figure*}[t!]
    \centering
    \includegraphics[width=1\textwidth]{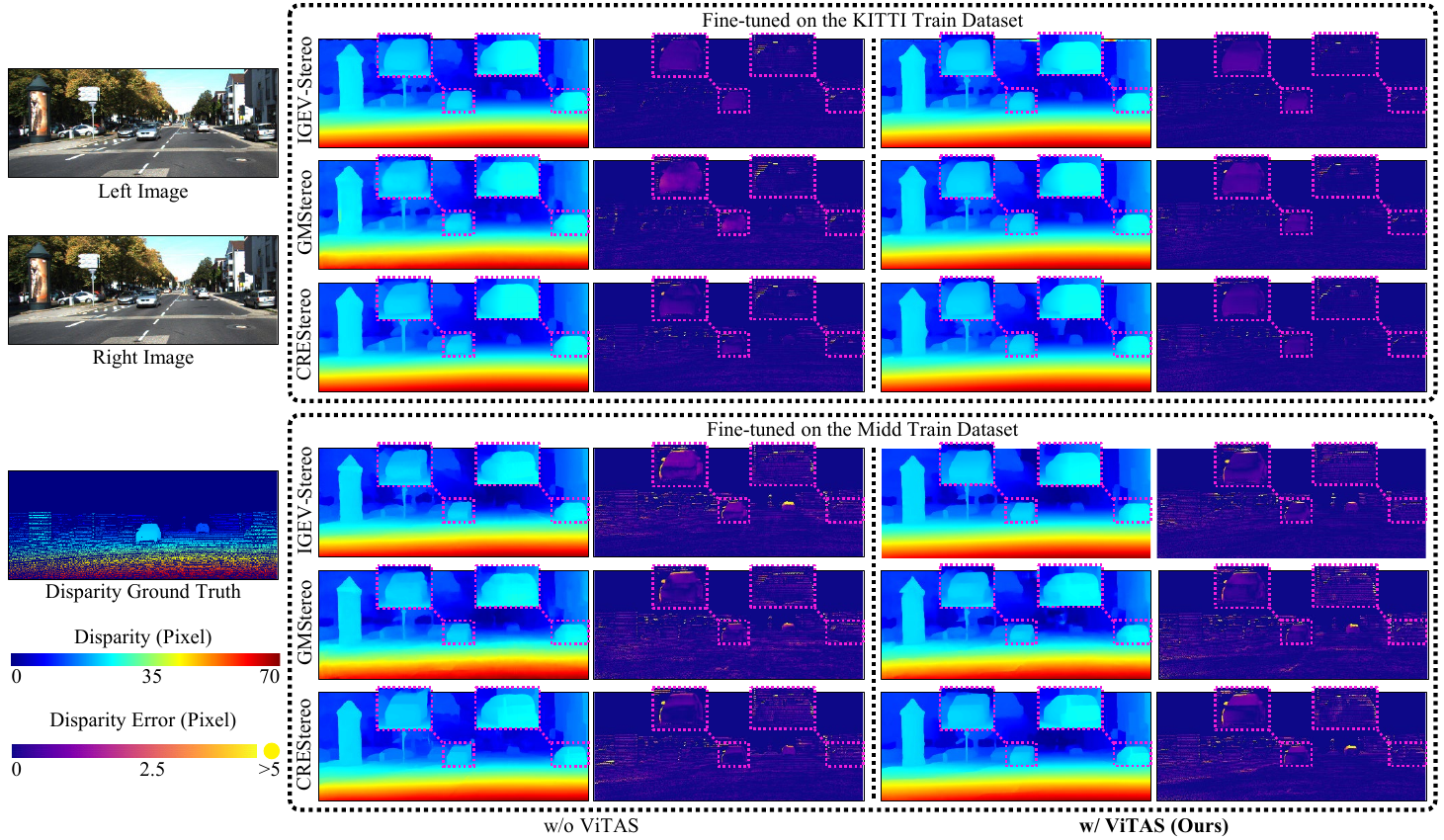}
    \centering
    \caption{Qualitative experimental results on the KITTI Eval dataset. Significantly improved regions are shown in pink dashed boxes.}
    \label{fig.general_kitti}
\end{figure*}

\begin{figure*}[t!]
    \centering
    \includegraphics[width=1\textwidth]{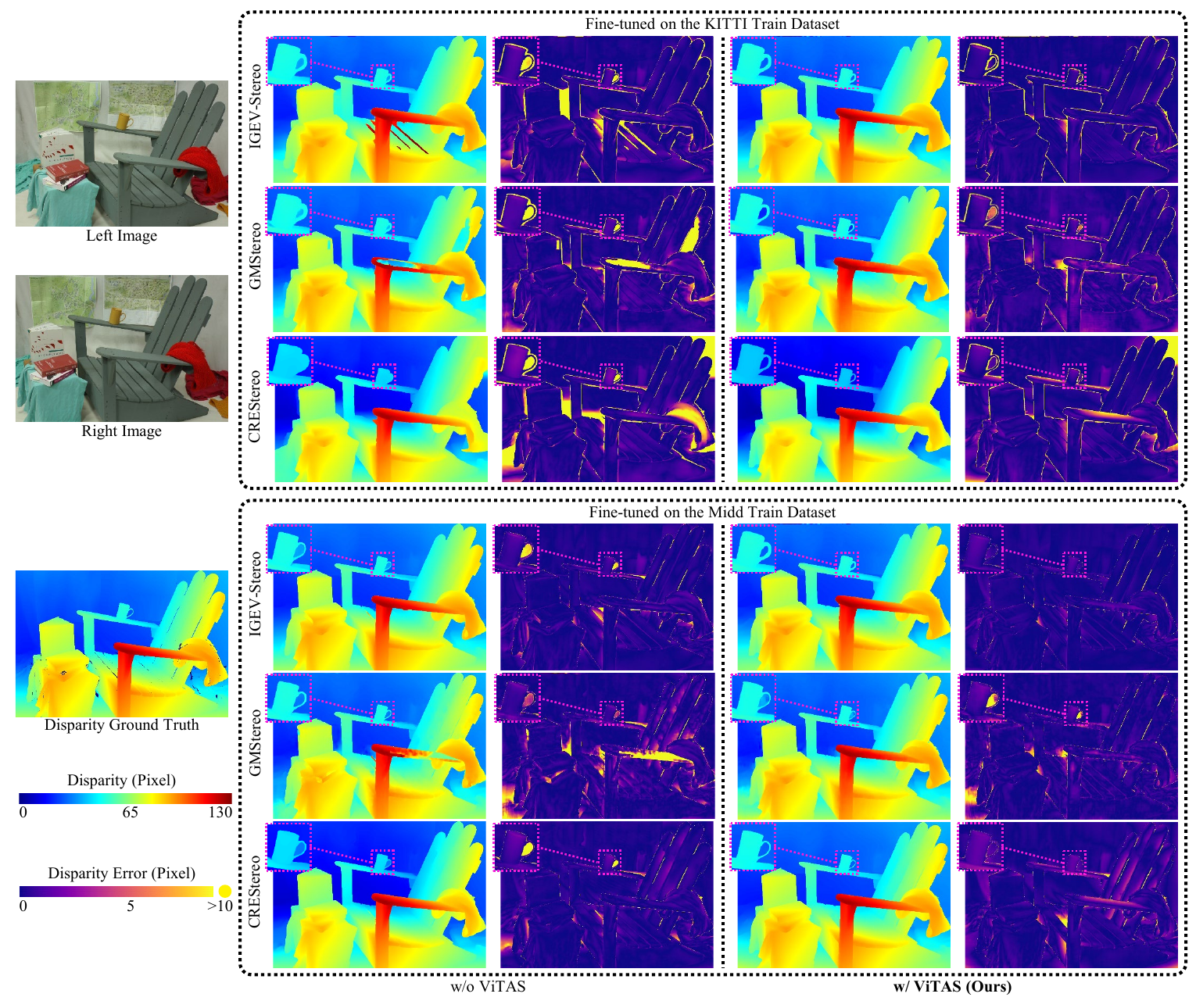}
    \centering
    \caption{Qualitative experimental results on the Midd Eval dataset. Significantly improved regions are shown in pink dashed boxes.}
    \label{fig.general_midd}
\end{figure*}

\begin{figure*}[t!]
    \centering
    \includegraphics[width=1\textwidth]{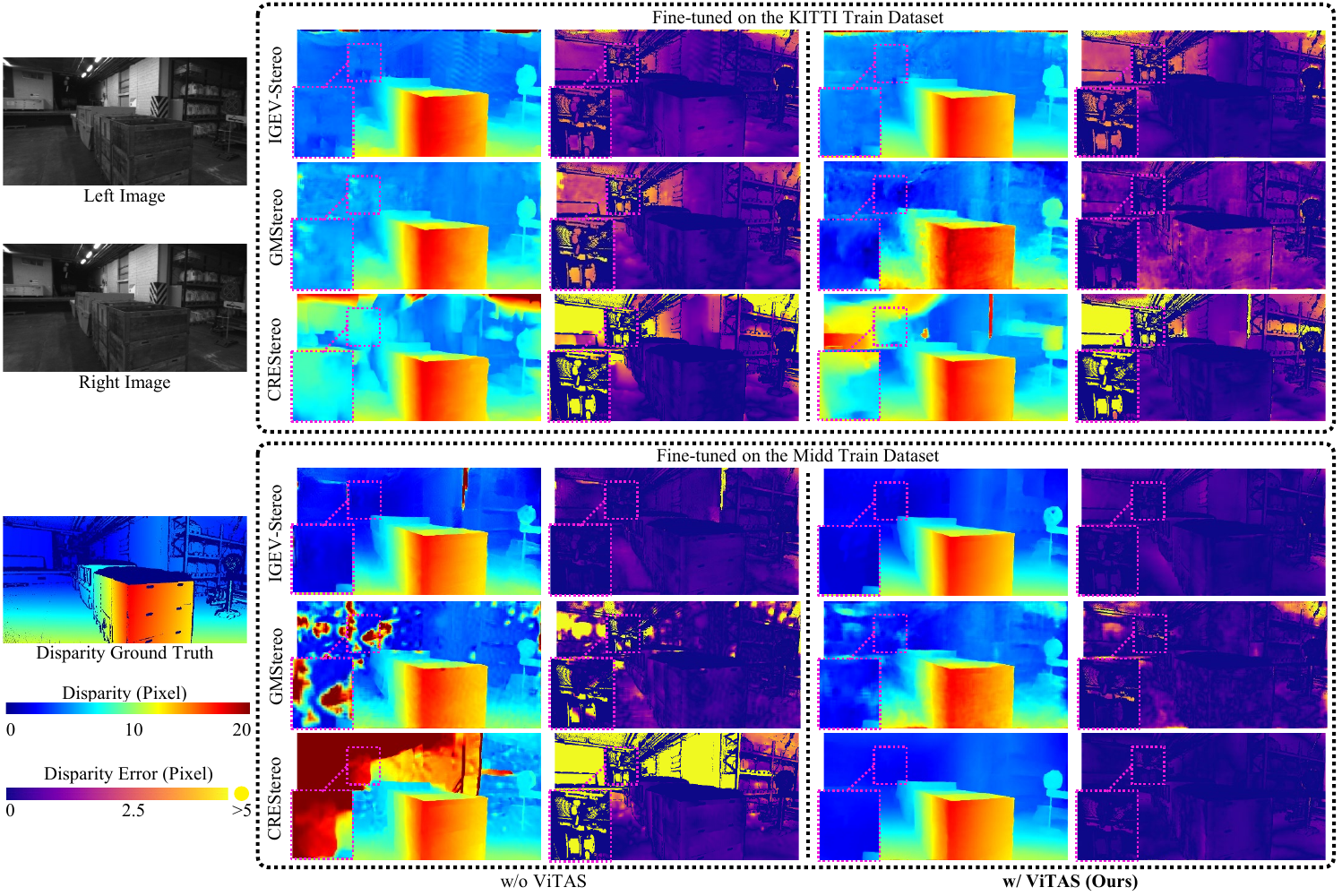}
    \centering
    \caption{Qualitative experimental results on the ETH3D dataset. Significantly improved regions are shown in pink dashed boxes.}
    \label{fig.general_eth3d}
\end{figure*}

\begin{table*}[t!]
    \settablefont
    % \parbox{12cm}
    \caption{Comparison among SoTA stereo matching networks without and with our proposed ViTAS applied. The best results are shown in bold type.}
    % \caption{Quantitative experimental results on the KITTI Eval, Midd Eval and ETH3D datasets. All the networks are initially pre-trained on the Sceneflow and vkitti datasets.}
    \centering
    \label{tab.general}
    {
    \setlength{\tabcolsep}{1.4mm}{
    \begin{tabular}{llcccccccc}
        \toprule
        \multirow{2}{*}{Network\hspace{2pt}} & \multirow{2}{*}[-0.4ex]{\makecell{Dataset for \vspace{-2pt} \\ \vspace{-1pt} Model Fine-tuning}} & \multirow{2}{*}[-0.4ex]{ViTAS\hspace{4pt}} & \multicolumn{2}{c}{KITTI Eval} & \multicolumn{2}{c}{Midd Eval} & \multicolumn{2}{c}{ETH3D} \\
        \cmidrule(r){4-5}
        \cmidrule(r){6-7}
        \cmidrule(r){8-9}
        \specialrule{0em}{-1pt}{-1pt}
        & & & EPE (px) & D1-all (\%) & EPE (px) & D1-all (\%) & EPE (px) & D1-all (\%)\\ 
        \hline
        \specialrule{0em}{1pt}{1pt}
        \multirow{4}{*}[-0.4ex]{IGEV-Stereo \cite{xu2023iterative}} & \multirow{2}{*}[-0.3ex]{KITTI Train} & w/o & 0.55 & 1.71 & 5.27 & 16.8 & 1.15 & 5.23 \\
        & & \textbf{w/} & \textbf{0.49} & \textbf{1.36} & \textbf{3.05} & \textbf{10.9} & \textbf{1.01} & \textbf{5.08} \\
        \cmidrule(r){2-9}
        & \multirow{2}{*}[-0.3ex]{Midd Train} & w/o & 1.07 & 5.27 & 2.14 & 10.8 & 4.20 & 5.49 \\
        & & \textbf{w/} & \textbf{0.87} & \textbf{3.45} & \textbf{1.34} & \textbf{6.00} & \textbf{2.65} & \textbf{3.68} \\
        \cmidrule(r){1-9}
        \multirow{4}{*}[-0.4ex]{GMStereo \cite{xu2023unifying}} & \multirow{2}{*}[-0.3ex]{KITTI Train} & w/o & 0.59 & 1.82 & 2.96 & 14.1 & 1.08 & 8.99 \\
        & & \textbf{w/} & \textbf{0.56} & \textbf{1.62} & \textbf{2.65} & \textbf{12.4} & \textbf{0.82} & \textbf{2.30} \\
        \cmidrule(r){2-9}
        & \multirow{2}{*}[-0.3ex]{Midd Train} & w/o & 1.14 & 6.14 & 2.65 & 13.3 & 1.72 & 10.1 \\
        &  & \textbf{w/} & \textbf{1.02} & \textbf{4.41} & \textbf{1.99} & \textbf{10.2} & \textbf{0.67} & \textbf{4.34} \\
        \cmidrule(r){1-9}
        \multirow{4}{*}[-0.4ex]{CREStereo \cite{li2022practical}} & \multirow{2}{*}[-0.3ex]{KITTI Train} & w/o & 0.70 & 2.32 & 5.20 & \textbf{16.1} & 3.45 & 18.8 \\
        &  & \textbf{w/} & \textbf{0.66} & \textbf{2.02} & \textbf{4.99} & {16.6} & \textbf{1.75} & \textbf{14.8}\\
        \cmidrule(r){2-9}
        & \multirow{2}{*}[-0.3ex]{Midd Train} & w/o & 1.18 & 6.24 & 3.98 & 14.4 & \textbf{28.6} & 35.0 \\
        & & \textbf{w/} & \textbf{1.07}& \textbf{4.91} & \textbf{2.59} & \textbf{12.7} & {29.1} & \textbf{25.9} \\
        \bottomrule  	
    \end{tabular}}}
\end{table*}

\begin{table*}[t!]
    \settablefont
    % \parbox{12cm}
    \caption{Quantitative experimental results of CroCo-Stereo \cite{weinzaepfel2023croco}.}
    \centering
    \label{tab.general_croco}
    {
    \setlength{\tabcolsep}{1.8mm}{
    \begin{tabular}{llccccccc}
        \toprule
        \multirow{2}{*}[-0.4ex]{Network\hspace{2pt}} & \multirow{2}{*}[-0.4ex]{\makecell{Dataset for \vspace{-2pt} \\ \vspace{-1pt} Model Fine-tuning} \hspace{5pt}} & \multicolumn{2}{c}{KITTI Eval} & \multicolumn{2}{c}{Midd Eval} & \multicolumn{2}{c}{ETH3D} \\
        \cmidrule(r){3-4}
        \cmidrule(r){5-6}
        \cmidrule(r){7-8}
        \specialrule{0em}{-1pt}{-1pt}
        & & EPE (px) & D1-all (\%) & EPE (px) & D1-all (\%) & EPE (px) & D1-all (\%)\\ 
        \specialrule{0em}{1pt}{1pt}
        \hline
        \specialrule{0em}{1pt}{1pt}
        \multirow{2}{*}{Original CroCo-Stereo \cite{weinzaepfel2023croco}} & {KITTI Train} & 1.16 & 6.17 & 8.80 & {25.1} & 51.5 & 97.4 \\
        & {Midd Train} & 6.70 & {35.0} & 2.04 & 9.90 & {50.7} & 88.0 \\
        \cmidrule(r){1-8}
        \multirow{2}{*}{Modified CroCo-Stereo} & {KITTI Train} & {0.86} & {3.26} & {8.15} & 25.4 & {50.7} & {89.1} \\
        & {Midd Train} & {4.05} & 44.9 & {1.87} & {8.82} & 62.9 & {85.2} \\
        \bottomrule  	
        \end{tabular}}}
\end{table*}

\begin{figure*}[t!]
    \centering
    \includegraphics[width=1\textwidth]{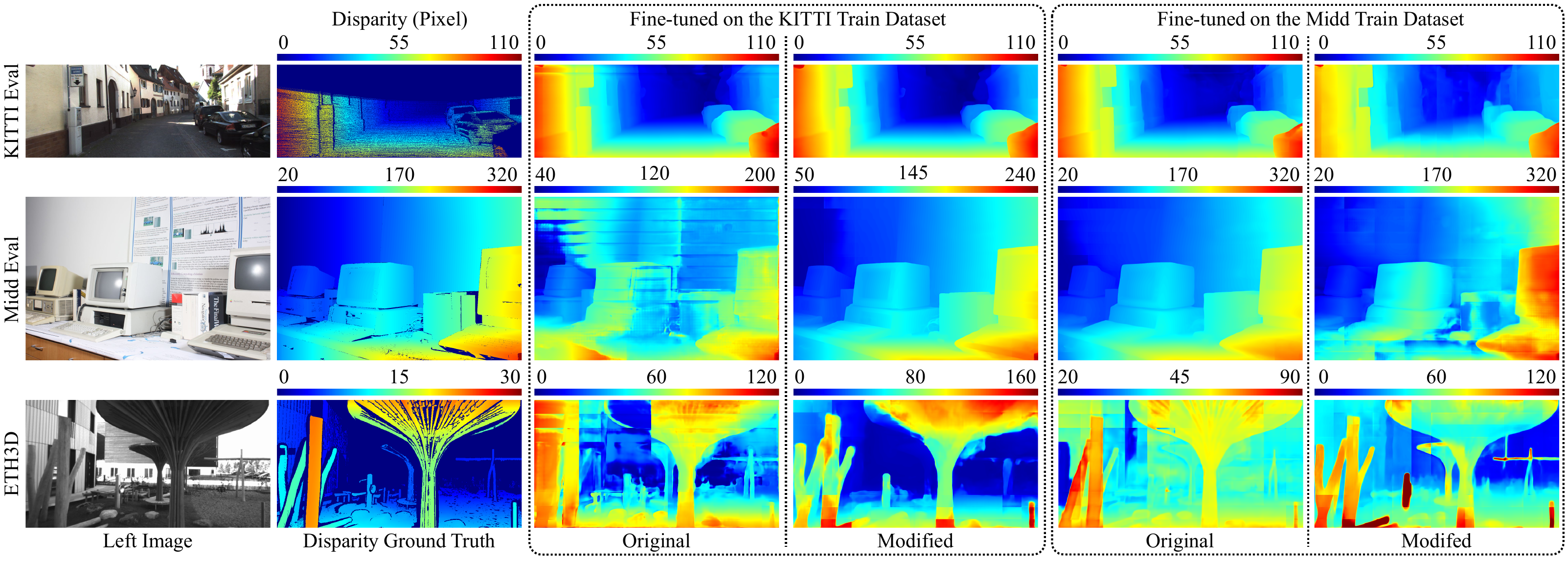}
    \centering
    \caption{Qualitative experimental results of CroCo-Stereo \cite{weinzaepfel2023croco}.}
    \label{fig.croco}
\end{figure*}

\subsection{Generalizability Evaluation}
\label{sec.cross}

\cl{
VFMs are renowned for their remarkable generalizability across diverse scenarios. Therefore, we conduct a series of additional experiments on three public, real-world datasets to further evaluate the generalizability of our proposed ViTAS, when combined with three SoTA cost volume-based stereo matching networks. The quantitative results are given in Table \ref{tab.general}, while the qualitative results on the KITTI Eval, Midd Eval, and ETH3D datasets are presented in Figs. \ref{fig.general_kitti}, \ref{fig.general_midd}, and \ref{fig.general_eth3d}, respectively. Our results suggest that leveraging ViTAS for visual feature extraction enables both IGEV-Stereo and GMStereo to consistently achieve superior performance across all evaluation metrics on each dataset. While CREStereo achieves comparable, albeit slightly less favorable, results in terms of D1-all (when trained on the KITTI Train dataset and evaluated on the Midd Eval dataset) and EPE (when trained on the Midd Train dataset and evaluated on the ETH3D dataset), it gains performance improvements in the remaining experiments. These extensive and comprehensive experimental results validate the compatibility of our proposed ViTAS as well as its effectiveness in adapting to new, unseen datasets.

To answer the question of whether cost volumes are becoming less critical or even expendable in SoTA stereo matching networks, we conduct an additional experiment with the recently published cost volume-free network, CroCo-Stereo, as detailed in Table \ref{tab.general_croco} and Fig. \ref{fig.croco}. In this experiment, we compare the performance of the original CroCo-Stereo against a modified version, where its ViT backbone is replaced with our utilized VFM. It can be observed in Table \ref{tab.general_croco} that both the original and modified CroCo-Stereo achieve accurate disparity estimation results on the KITTI Eval and Midd Eval datasets after being fine-tuned on the KITTI Train and Midd Train datasets, respectively. However, their accuracy in disparity estimation notably decreases on the other two evaluation datasets, with significantly higher EPE and D1-all observed on the ETH3D dataset. The results presented in Fig. \ref{fig.croco} emphasize a significant issue with both networks: scale ambiguity. This is evident in the \textbf{relatively smaller disparity estimation results} on the Midd Eval dataset after fine-tuning on the KITTI Train dataset, as well as the \textbf{relatively larger disparity estimation results} on the ETH3D dataset. Hence, we argue that cost volume remains essential, particularly when aiming for generalizable stereo matching based on VFMs.
}

\section{Conclusion and Future Work}
\label{sec.conc}

\cl{
This article introduced ViTAS, a pioneering research effort to fully exploit the general-purpose VFM features for stereo matching. Our study has yielded several key findings: (1) the prevalent use of CNN-based feature extractors in existing stereo matching networks has been identified as a critical bottleneck, limiting these networks from achieving higher levels of stereo matching accuracy; (2) a pre-trained VFM combined with an appropriate adapter, demonstrates superior performance in terms of both accuracy and generalizability, when compared to conventional CNN-based backbones; (3) merely aggregating stereo contextual information via the cross-attention mechanism falls short in addressing the scale ambiguity problem, underscoring the indispensable role of cost volumes in developing generalizable stereo matching networks. Our ViTAStereo sets a new standard in performance on the KITTI Stereo datasets, establishing itself as the state-of-the-art in this field. While the contributions of this study are significant, it is noted that the CAM utilized herein remains unchanged and still requires substantial computational and memory resources. Therefore, we intend to investigate more efficient strategies for the aggregation of stereo contextual information in the near future. 
}

\normalem
\bibliographystyle{IEEEtran}

\end{document}

%% file: packages.tex
\usepackage[table,xcdraw,dvipsnames]{xcolor}
\usepackage{scalerel}
\usepackage{tikz}
\usetikzlibrary{svg.path}

\definecolor{orcidlogocol}{HTML}{A6CE39}
\tikzset{
  orcidlogo/.pic={
    \fill[orcidlogocol] svg{M256,128c0,70.7-57.3,128-128,128C57.3,256,0,198.7,0,128C0,57.3,57.3,0,128,0C198.7,0,256,57.3,256,128z};
    \fill[white] svg{M86.3,186.2H70.9V79.1h15.4v48.4V186.2z}
                 svg{M108.9,79.1h41.6c39.6,0,57,28.3,57,53.6c0,27.5-21.5,53.6-56.8,53.6h-41.8V79.1z M124.3,172.4h24.5c34.9,0,42.9-26.5,42.9-39.7c0-21.5-13.7-39.7-43.7-39.7h-23.7V172.4z}
                 svg{M88.7,56.8c0,5.5-4.5,10.1-10.1,10.1c-5.6,0-10.1-4.6-10.1-10.1c0-5.6,4.5-10.1,10.1-10.1C84.2,46.7,88.7,51.3,88.7,56.8z};
  }
}
\newcommand\orcidicon[1]{\href{https://orcid.org/#1}{\mbox{\scalerel*{
\begin{tikzpicture}[yscale=-1,transform shape]
\pic{orcidlogo};
\end{tikzpicture}
}{|}}}}

%% file: main.bbl
\begin{thebibliography}{10}
	\providecommand{\url}[1]{#1}
	\csname url@samestyle\endcsname
	\providecommand{\newblock}{\relax}
	\providecommand{\bibinfo}[2]{#2}
	\providecommand{\BIBentrySTDinterwordspacing}{\spaceskip=0pt\relax}
	\providecommand{\BIBentryALTinterwordstretchfactor}{4}
	\providecommand{\BIBentryALTinterwordspacing}{\spaceskip=\fontdimen2\font plus
		\BIBentryALTinterwordstretchfactor\fontdimen3\font minus
		\fontdimen4\font\relax}
	\providecommand{\BIBforeignlanguage}[2]{{%
			\expandafter\ifx\csname l@#1\endcsname\relax
			\typeout{** WARNING: IEEEtran.bst: No hyphenation pattern has been}%
			\typeout{** loaded for the language `#1'. Using the pattern for}%
			\typeout{** the default language instead.}%
			\else
			\language=\csname l@#1\endcsname
			\fi
			#2}}
	\providecommand{\BIBdecl}{\relax}
	\BIBdecl
	
	\bibitem{kirillov2023segment}
	A.~Kirillov \emph{et~al.}, ``Segment anything,'' in \emph{Proceedings of the
		IEEE/CVF International Conference on Computer Vision (ICCV)}, 2023, pp.
	4015--4026.
	
	\bibitem{oquab2023dinov2}
	M.~Oquab \emph{et~al.}, ``{DINOv2}: Learning robust visual features without
	supervision,'' \emph{arXiv preprint arXiv:2304.07193}, 2023.
	
	\bibitem{yang2024depth}
	L.~Yang \emph{et~al.}, ``Depth anything: Unleashing the power of large-scale
	unlabeled data,'' \emph{arXiv preprint arXiv:2401.10891}, 2024.
	
	\bibitem{guo2019group}
	X.~Guo \emph{et~al.}, ``Group-wise correlation stereo network,'' in
	\emph{Proceedings of the IEEE/CVF Conference on Computer Vision and Pattern
		Recognition (CVPR)}, 2019, pp. 3273--3282.
	
	\bibitem{xu2023iterative}
	G.~Xu \emph{et~al.}, ``Iterative geometry encoding volume for stereo
	matching,'' in \emph{Proceedings of the IEEE/CVF Conference on Computer
		Vision and Pattern Recognition (CVPR)}, 2023, pp. 21\,919--21\,928.
	
	\bibitem{xu2023unifying}
	H.~Xu \emph{et~al.}, ``Unifying flow, stereo and depth estimation,'' \emph{IEEE
		Transactions on Pattern Analysis and Machine Intelligence}, vol.~45, no.~11,
	pp. 13\,941--13\,958, 2023.
	
	\bibitem{he2016deep}
	K.~He \emph{et~al.}, ``Deep residual learning for image recognition,'' in
	\emph{Proceedings of the IEEE Conference on Computer Vision and Pattern
		Recognition (CVPR)}, 2016, pp. 770--778.
	
	\bibitem{sandler2018mobilenetv2}
	M.~Sandler and pthers, ``{MobileNetV2}: Inverted residuals and linear
	bottlenecks,'' in \emph{Proceedings of the IEEE Conference on Computer Vision
		and Pattern Recognition (CVPR)}, 2018, pp. 4510--4520.
	
	\bibitem{li2021benchmarking}
	Y.~Li \emph{et~al.}, ``Benchmarking detection transfer learning with vision
	{Transformers},'' \emph{arXiv preprint arXiv:2111.11429}, 2021.
	
	\bibitem{li2022exploring}
	Y.~Li \emph{et~al.}, ``Exploring plain vision {Transformer} backbones for
	object detection,'' in \emph{European Conference on Computer Vision
		(ECCV)}.\hskip 1em plus 0.5em minus 0.4em\relax Springer, 2022, pp. 280--296.
	
	\bibitem{chen2023vision}
	Z.~Chen \emph{et~al.}, ``Vision {Transformer} adapter for dense predictions,''
	in \emph{International Conference on Learning Representations (ICLR)}, 2023.
	
	\bibitem{li2022practical}
	J.~Li \emph{et~al.}, ``Practical stereo matching via cascaded recurrent network
	with adaptive correlation,'' in \emph{Proceedings of the IEEE/CVF Conference
		on Computer Vision and Pattern Recognition (CVPR)}, 2022, pp.
	16\,263--16\,272.
	
	\bibitem{guo2022context}
	W.~Guo \emph{et~al.}, ``Context-enhanced stereo {Transformer},'' in
	\emph{European Conference on Computer Vision (ECCV)}.\hskip 1em plus 0.5em
	minus 0.4em\relax Springer, 2022, pp. 263--279.
	
	\bibitem{liu2024global}
	Z.~Liu \emph{et~al.}, ``Global occlusion-aware {Transformer} for robust stereo
	matching,'' in \emph{Proceedings of the IEEE/CVF Winter Conference on
		Applications of Computer Vision (WACV)}, 2024, pp. 3535--3544.
	
	\bibitem{weinzaepfel2023croco}
	P.~Weinzaepfel \emph{et~al.}, ``{CroCo} v2: Improved cross-view completion
	pre-training for stereo matching and optical flow,'' in \emph{Proceedings of
		the IEEE/CVF International Conference on Computer Vision (ICCV)}, 2023, pp.
	17\,969--17\,980.
	
	\bibitem{weinzaepfel2022croco}
	P.~Weinzaepfel \emph{et~al.}, ``{CroCo}: Self-supervised pre-training for {3D}
	vision tasks by cross-view completion,'' \emph{Advances in Neural Information
		Processing Systems (NeurIPS)}, vol.~35, pp. 3502--3516, 2022.
	
	\bibitem{ranftl2021vision}
	R.~Ranftl \emph{et~al.}, ``Vision {Transformers} for dense prediction,'' in
	\emph{Proceedings of the IEEE/CVF International Conference on Computer Vision
		(ICCV)}, 2021, pp. 12\,179--12\,188.
	
	\bibitem{eigen2014depth}
	D.~Eigen \emph{et~al.}, ``Depth map prediction from a single image using a
	multi-scale deep network,'' \emph{Advances in Neural Information Processing
		Systems (NeurIPS)}, vol.~27, 2014.
	
	\bibitem{lipson2021raft}
	L.~Lipson \emph{et~al.}, ``{RAFT-Stereo}: Multilevel recurrent field transforms
	for stereo matching,'' in \emph{2021 International Conference on 3D Vision
		(3DV)}.\hskip 1em plus 0.5em minus 0.4em\relax IEEE, 2021, pp. 218--227.
	
	\bibitem{shen2021cfnet}
	Z.~Shen \emph{et~al.}, ``{CFNet}: Cascade and fused cost volume for robust
	stereo matching,'' in \emph{Proceedings of the IEEE/CVF Conference on
		Computer Vision and Pattern Recognition (CVPR)}, 2021, pp. 13\,906--13\,915.
	
	\bibitem{zhou2022self}
	Z.~Zhou and Q.~Dong, ``Self-distilled feature aggregation for self-supervised
	monocular depth estimation,'' in \emph{European Conference on Computer Vision
		(ECCV)}.\hskip 1em plus 0.5em minus 0.4em\relax Springer, 2022, pp. 709--726.
	
	\bibitem{zhao2022semantic}
	Y.~Zhao \emph{et~al.}, ``Semantic-aligned fusion {Transformer} for one-shot
	object detection,'' in \emph{Proceedings of the IEEE/CVF Conference on
		Computer Vision and Pattern Recognition (CVPR)}, 2022, pp. 7601--7611.
	
	\bibitem{geiger2012we}
	A.~Geiger \emph{et~al.}, ``Are we ready for autonomous driving? the kitti
	vision benchmark suite,'' in \emph{2012 IEEE Conference on Computer Vision
		and Pattern Recognition (CVPR)}.\hskip 1em plus 0.5em minus 0.4em\relax IEEE,
	2012, pp. 3354--3361.
	
	\bibitem{xie2021propagate}
	Z.~Xie \emph{et~al.}, ``Propagate yourself: Exploring pixel-level consistency
	for unsupervised visual representation learning,'' in \emph{Proceedings of
		the IEEE/CVF Conference on Computer Vision and Pattern Recognition (CVPR)},
	2021, pp. 16\,684--16\,693.
	
	\bibitem{wang2021dense}
	X.~Wang \emph{et~al.}, ``Dense contrastive learning for self-supervised visual
	pre-training,'' in \emph{Proceedings of the IEEE/CVF Conference on Computer
		Vision and Pattern Recognition (CVPR)}, 2021, pp. 3024--3033.
	
	\bibitem{sun2021loftr}
	J.~Sun \emph{et~al.}, ``{LoFTR}: Detector-free local feature matching with
	{Transformers},'' in \emph{Proceedings of the IEEE/CVF Conference on Computer
		Vision and Pattern Recognition (CVPR)}, 2021, pp. 8922--8931.
	
	\bibitem{deng2009imagenet}
	J.~Deng \emph{et~al.}, ``{ImageNet}: A large-scale hierarchical image
	database,'' in \emph{2009 IEEE Conference on Computer Vision and Pattern
		Recognition (CVPR)}.\hskip 1em plus 0.5em minus 0.4em\relax IEEE, 2009, pp.
	248--255.
	
	\bibitem{wang2022pvt}
	W.~Wang \emph{et~al.}, ``{PVT} v2: Improved baselines with pyramid vision
	{Transformer},'' \emph{Computational Visual Media}, vol.~8, no.~3, pp.
	415--424, 2022.
	
	\bibitem{quan2023centralized}
	Y.~Quan \emph{et~al.}, ``Centralized feature pyramid for object detection,''
	\emph{IEEE Transactions on Image Processing}, vol.~32, pp. 4341--4354, 2023.
	
	\bibitem{ma2022multiview}
	Z.~Ma \emph{et~al.}, ``Multiview stereo with cascaded epipolar raft,'' in
	\emph{European Conference on Computer Vision (ECCV)}.\hskip 1em plus 0.5em
	minus 0.4em\relax Springer, 2022, pp. 734--750.
	
	\bibitem{xu2023accurate}
	G.~Xu \emph{et~al.}, ``Accurate and efficient stereo matching via attention
	concatenation volume,'' \emph{IEEE Transactions on Pattern Analysis and
		Machine Intelligence}, pp. 1--13, 2023.
	
	\bibitem{dosovitskiy2020image}
	A.~Dosovitskiy \emph{et~al.}, ``An image is worth 16x16 words: {Transformers}
	for image recognition at scale,'' in \emph{International Conference on
		Learning Representations (ICLR)}, 2021.
	
	\bibitem{park2022vision}
	N.~Park and S.~Kim, ``How do vision {Transformers} work?'' in
	\emph{International Conference on Learning Representations (ICLR)}, 2021.
	
	\bibitem{fang2023unleashing}
	Y.~Fang \emph{et~al.}, ``Unleashing vanilla vision {Transformer} with masked
	image modeling for object detection,'' in \emph{Proceedings of the IEEE/CVF
		International Conference on Computer Vision (ICCV)}, 2023, pp. 6244--6253.
	
	\bibitem{wang2023cross}
	G.~Wang \emph{et~al.}, ``Cross-level attentive feature aggregation for change
	detection,'' \emph{IEEE Transactions on Circuits and Systems for Video
		Technology}, 2023, doi:{\color{black}
		\href{https://ieeexplore.ieee.org/abstract/document/10364762}{10.1109/TCSVT.2023.3344092}}.
	
	\bibitem{fan2020sne}
	R.~Fan \emph{et~al.}, ``{SNE-RoadSeg}: Incorporating surface normal information
	into semantic segmentation for accurate freespace detection,'' in
	\emph{European Conference on Computer Vision (ECCV)}.\hskip 1em plus 0.5em
	minus 0.4em\relax Springer, 2020, pp. 340--356.
	
	\bibitem{chang2020attention}
	J.-R. Chang \emph{et~al.}, ``Attention-aware feature aggregation for real-time
	stereo matching on edge devices,'' in \emph{Proceedings of the Asian
		Conference on Computer Vision (ACCV)}, 2020.
	
	\bibitem{liu2023learning}
	W.~Liu \emph{et~al.}, ``Learning to upsample by learning to sample,'' in
	\emph{Proceedings of the IEEE/CVF International Conference on Computer Vision
		(ICCV)}, 2023, pp. 6027--6037.
	
	\bibitem{liu2023devil}
	Y.~Liu \emph{et~al.}, ``The devil is in the upsampling: Architectural decisions
	made simpler for denoising with deep image prior,'' in \emph{Proceedings of
		the IEEE/CVF International Conference on Computer Vision (ICCV)}, 2023, pp.
	12\,408--12\,417.
	
	\bibitem{hu2018squeeze}
	J.~Hu \emph{et~al.}, ``Squeeze-and-excitation networks,'' in \emph{Proceedings
		of the IEEE Conference on Computer Vision and Pattern Recognition (CVPR)},
	2018, pp. 7132--7141.
	
	\bibitem{gao2022doubly}
	L.~Gao \emph{et~al.}, ``Doubly-fused {ViT}: Fuse information from vision
	{Transformer} doubly with local representation,'' in \emph{European
		Conference on Computer Vision (ECCV)}.\hskip 1em plus 0.5em minus 0.4em\relax
	Springer, 2022, pp. 744--761.
	
	\bibitem{su2022chitransformer}
	Q.~Su and S.~Ji, ``{ChiTransformer}: Towards reliable stereo from cues,'' in
	\emph{Proceedings of the IEEE/CVF Conference on Computer Vision and Pattern
		Recognition (CVPR)}, 2022, pp. 1939--1949.
	
	\bibitem{mayer2016large}
	N.~Mayer \emph{et~al.}, ``A large dataset to train convolutional networks for
	disparity, optical flow, and scene flow estimation,'' in \emph{Proceedings of
		the IEEE Conference on Computer Vision and Pattern Recognition (CVPR)}, 2016,
	pp. 4040--4048.
	
	\bibitem{menze2015object}
	M.~Menze and A.~Geiger, ``Object scene flow for autonomous vehicles,'' in
	\emph{Proceedings of the IEEE Conference on Computer Vision and Pattern
		Recognition (CVPR)}, 2015, pp. 3061--3070.
	
	\bibitem{scharstein2007learning}
	D.~Scharstein and C.~Pal, ``Learning conditional random fields for stereo,'' in
	\emph{Proceedings of the IEEE Conference on Computer Vision and Pattern
		Recognition (CVPR)}.\hskip 1em plus 0.5em minus 0.4em\relax IEEE, 2007, pp.
	1--8.
	
	\bibitem{hirschmuller2007evaluation}
	H.~Hirschmuller and D.~Scharstein, ``Evaluation of cost functions for stereo
	matching,'' in \emph{2007 IEEE Conference on Computer Vision and Pattern
		Recognition (CVPR)}.\hskip 1em plus 0.5em minus 0.4em\relax IEEE, 2007, pp.
	1--8.
	
	\bibitem{scharstein2014high}
	D.~Scharstein \emph{et~al.}, ``High-resolution stereo datasets with
	subpixel-accurate ground truth,'' in \emph{Pattern Recognition: 36th German
		Conference (GCPR)}.\hskip 1em plus 0.5em minus 0.4em\relax Springer, 2014,
	pp. 31--42.
	
	\bibitem{schops2017multi}
	T.~Schops \emph{et~al.}, ``A multi-view stereo benchmark with high-resolution
	images and multi-camera videos,'' in \emph{Proceedings of the IEEE Conference
		on Computer Vision and Pattern Recognition (CVPR)}, 2017, pp. 3260--3269.
	
	\bibitem{cheng2020hierarchical}
	X.~Cheng \emph{et~al.}, ``Hierarchical neural architecture search for deep
	stereo matching,'' \emph{Advances in Neural Information Processing Systems
		(NeurIPS)}, vol.~33, pp. 22\,158--22\,169, 2020.
	
	\bibitem{tankovich2021hitnet}
	V.~Tankovich \emph{et~al.}, ``{HITnet}: Hierarchical iterative tile refinement
	network for real-time stereo matching,'' in \emph{Proceedings of the IEEE/CVF
		Conference on Computer Vision and Pattern Recognition (CVPR)}, 2021, pp.
	14\,362--14\,372.
	
	\bibitem{xu2022attention}
	G.~Xu \emph{et~al.}, ``Attention concatenation volume for accurate and
	efficient stereo matching,'' in \emph{Proceedings of the IEEE/CVF Conference
		on Computer Vision and Pattern Recognition (CVPR)}, 2022, pp.
	12\,981--12\,990.
	
	\bibitem{shen2022pcw}
	Z.~Shen \emph{et~al.}, ``{PCW-Net}: Pyramid combination and warping cost volume
	for stereo matching,'' in \emph{European Conference on Computer Vision
		(ECCV)}.\hskip 1em plus 0.5em minus 0.4em\relax Springer, 2022, pp. 280--297.
	
	\bibitem{shen2023digging}
	Z.~Shen \emph{et~al.}, ``Digging into uncertainty-based pseudo-label for robust
	stereo matching,'' \emph{IEEE Transactions on Pattern Analysis and Machine
		Intelligence}, vol.~45, no.~12, pp. 14\,301--14\,320, 2023.
	
	\bibitem{kwon2023image}
	O.-H. Kwon and E.~Zell, ``Image-coupled volume propagation for stereo
	matching,'' in \emph{2023 IEEE International Conference on Image Processing
		(ICIP)}.\hskip 1em plus 0.5em minus 0.4em\relax IEEE, 2023, pp. 2510--2514.
	
	\bibitem{feng2023mc}
	M.~Feng \emph{et~al.}, ``{MC-Stereo}: Multi-peak lookup and cascade search
	range for stereo matching,'' \emph{arXiv preprint arXiv:2311.02340}, 2023.
	
	\bibitem{guan2024neural}
	T.~Guan \emph{et~al.}, ``Neural {Markov} random field for stereo matching,''
	\emph{arXiv preprint arXiv:2403.11193}, 2024.
	
	\bibitem{guo2023openstereo}
	X.~Guo \emph{et~al.}, ``{OpenStereo}: A comprehensive benchmark for stereo
	matching and strong baseline,'' \emph{arXiv preprint arXiv:2312.00343}, 2023.
	
\end{thebibliography}
